\documentclass[letterpaper, 10 pt]{ieeeconf}  % Comment this line out if you need a4paper

\IEEEoverridecommandlockouts                              % This command is only needed if 
\usepackage{subcaption}

\usepackage{graphicx} % for pdf, bitmapped graphics files
\usepackage{tikz}
\usepackage{pgfplots}
\usepackage{amsmath} % assumes amsmath package installed
\usepackage{amssymb}  % assumes amsmath package installed
\usepackage{import}
\usepackage{dsfont}
\usepackage{hyperref}
\usepackage{gensymb}
\usepackage{booktabs}

\title{\LARGE \bf
Long-Term Occupancy Grid Prediction Using\\Recurrent Neural Networks
}

\author{Marcel Schreiber$^{*}$, Stefan Hoermann$^{*}$, and Klaus Dietmayer
	\thanks{The authors are with Institute of Measurement, Control, and Microtechnology, Ulm University, Germany 
		{\tt\footnotesize \{first.last\}@uni-ulm.de}}% <-this % stops a space
	\thanks{$^{*}$ indicates equal contribution}%
}
\begin{document}

\bstctlcite{IEEEexample:BSTcontrol} 

\def\cred{\textcolor{red}}
\def\cblue{\textcolor{blue}}
\def\cgreen{\textcolor{green}}
\newcommand\copyrighttext{%
	\footnotesize This work has been submitted to the IEEE for possible publication. Copyright may be transferred without notice, after which this version may no longer be accessible.}%
\newcommand\copyrightnotice{%
	\begin{tikzpicture}[remember picture,overlay]%
	\node[anchor=south,yshift=10pt] at (current page.south) {\fbox{\parbox{\dimexpr\textwidth}{\copyrighttext}}};%
	\end{tikzpicture}%
	\vspace{-10pt}%
}

\maketitle
\copyrightnotice%
\thispagestyle{empty}
\pagestyle{empty}

%%%%%%%%%%%%%%%%%%%%%%%%%%%%%%%%%%%%%%%%%%%%%%%%%%%%%%%%%%%%%%%%%%%%%%%%%%%%%%%%
\begin{abstract}
We tackle the long\hbox{-}term prediction of scene evolution in a complex downtown scenario for automated driving based on Lidar grid fusion and recurrent neural networks (RNNs).
A bird's eye view of the scene, including occupancy and velocity, is fed as a sequence to a RNN which is trained to predict future occupancy.
The nature of prediction allows generation of multiple hours of training data without the need of manual labeling. 
Thus, the training strategy and loss function are designed for long sequences of real-world data (unbalanced, continuously changing situations, false labels, etc.).
The deep CNN architecture comprises convolutional long short\hbox{-}term memories (ConvLSTMs) to separate static from dynamic regions and to predict dynamic objects in future frames.
Novel recurrent skip connections show the ability to predict small occluded objects, i.e. pedestrians, and occluded static regions.
Spatio-temporal correlations between grid cells are exploited to predict multimodal future paths and interactions between objects.
Experiments also quantify improvements to our previous network, a Monte Carlo approach, and literature.

\end{abstract}
%

%%%%%%%%%%%%%%%%%%%%%%%%%%%%%%%%%%%%%%%%%%%%%%%%%%%%%%%%%%%%%%%%%%%%%%%%%%%%%%%%
\section{Introduction}
A human driver has the ability to estimate behavior of other road users and predict their motion for the next few seconds.
This enables foresighted driving and compensates a relatively slow reaction compared to automated vehicles.
For an automated system, in contrast, this long-term prediction of the scene evolution is one of the hardest tasks. 
We tackle this problem to make a long-term prediction of the future vehicle environment in a complex downtown scenario with different road users like pedestrians, cyclists and vehicles.
The complex and diverse behavior of these road users is hard to describe with handcrafted models.
Furthermore, an object's behavior not only depends on its own current dynamics, but also on the environment.
Thus, it seems reasonable to design a long-term prediction to be able to exploit information of the current object dynamics, the context of the environment, and also the past scene evolution.
We achieve such a design using the dynamic occupancy grid map (DOGMa) as input to a recurrent neural network.

\begin{figure}[t]
	\centering
	\includegraphics[width=\columnwidth]{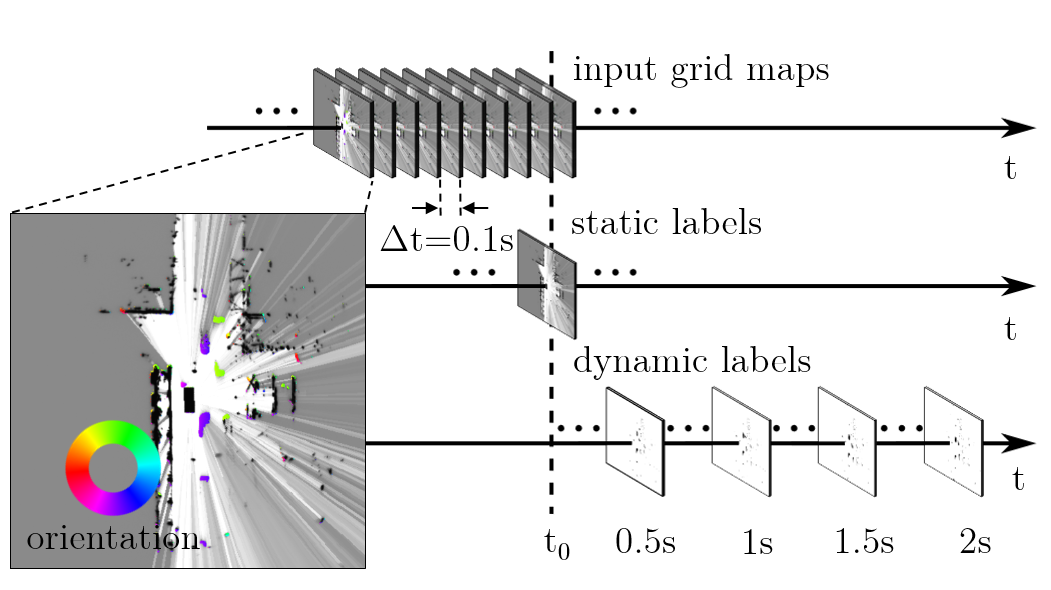}
	\caption{
		Input data comprises a sequence of dynamic occupancy grid maps which appear as a bird's eye view containing velocity and occupancy estimates.
		Also static and dynamic labels are stored as continuous sequences. 
		This allows varying time intervals (future and past) for training.}
	\label{fig:datastructure2}
\end{figure}

The DOGMa as shown in Fig. \ref{fig:datastructure2} is an extended version of classical grid maps.
It contains occupancy and velocity estimates in each cell, obtained by a Monte Carlo approach \cite{DBLP:journals/corr/NussRTYKMGD16}.  
Based on our previous work \cite{DBLP:journals/corr/HoermannBD17}, we use DOGMas as input of a neural network, containing convolutional and recurrent layers to predict the future scene in form of occupancy grid maps.
A key component in this work is the use of convolutional long short-term memories (ConvLSTMs) \cite{DBLP:journals/corr/ShiCWYWW15} to exploit spatial context and capture temporal correlations between time steps.
The recurrent network part is built of an Encoder\hbox{-}LSTM to perform a sequential filtering of the input scene and a Decoder\hbox{-}LSTM to produce predictions for several time horizons.
In addition, the developed network architecture is extended with a novel recurrent skip architecture, which shows beneficial characteristics, when input data is missing due to occlusions in the input grid map.
Furthermore, a training strategy for recurrent neural networks is proposed with special attention on long training data sequences.
The strategy is based on random selected sequence entry points and freely selectable subsequence length, e.g. relevant past and desired prediction horizon.
The remainder of the paper is structured as follows: Section \ref{sec:related_work} gives a review of related work. 
The input data is described in Section \ref{sec:network}, moreover the network architecture of the Grid Predictor Model and the extension with a novel recurrent skip architecture are introduced.
Section \ref{sec:dataset} gives a brief overview of the dataset.
In addition, the spatial balancing loss, the data structure, and the training process are introduced.
The experiments comparing the results with our previous approach, a Monte Carlo alternative, and results from related work are presented in Section \ref{sec:Eval}.
Also experiments with occlusion, multimodality, and interactions are presented.
The paper is concluded in Section \ref{sec:conclusions}.
\section{Related Work}\label{sec:related_work}
The present work is a continuation of our previous work for occupancy grid prediction \cite{DBLP:journals/corr/HoermannBD17}.
A main drawback is the assumption of independent cells in the DOGMa creation, which is based on Bayesian sequential filtering.
In the presented approach sequential filtering is also performed using recurrent network structures.
Two times sequential filtering (input data and predictor network) is motivated by first estimating the dynamics in the scene, and in a second stage exploiting spatio-temporal correlations between cells, i.e. interacting objects occupying multiple cells.

Due to the image-like data structure of DOGMas, the task of predicting future occupancy grid maps, based on the sequence of past DOGMas, is comparable to a video prediction problem. 
So, the recurrent network part is inspired by the Encoder\hbox{-}Decoder framework introduced by Srivastava et~al. in \cite{DBLP:journals/corr/SrivastavaMS15} based on \cite{DBLP:journals/corr/SutskeverVL14} to predict future video frames using fully connected LSTMs (FC\hbox{-}LSTMs).
We use the sequence-to-sequence mapping ability to train the encoder to observe the input sequence and the decoder to produce multiple future grid maps. 
LSTMs were first introduced by Hochreiter and Schmidhuber \cite{Hochreiter97longshort-term}, extended in \cite{GersLearningtoforget}, \cite{Gers:2000:RNT:870462} and overcome the problems of learning long\hbox{-}term dependencies, described in \cite{Pascanu:2013:DTR:3042817.3043083}.
To reduce parameters, we use convolutional LSTMs (ConvLSTMs), which were introduced in \cite{DBLP:journals/corr/ShiCWYWW15} and outperform the FC\hbox{-}LSTMs in capturing spatio\hbox{-}temporal relationships.

Literature covers alternative neural network based approaches for grid based scene prediction, mostly differing in the used input data and application.
E.g., the work of \cite{2018arXiv180206338P} uses rectangular grids of $5.0\times0.87\,\mathrm{m}^2$ to predict vehicle trajectories in a highway scenario using an Encoder\hbox{-}Decoder LSTM architecture.
In contrast to our work, detected objects are placed in the grid, instead of raw measurements.
We also use a finer $15\times15\,\mathrm{cm}^2$ grid and focus on an urban shared space scenario.
From an environment model and application perspective, the Deep Tracking series \cite{doi:10.1177/0278364917710543, DBLP:journals/corr/OndruskaP16, DBLP:journals/corr/OndruskaDWP16} is closely related work.
The basic idea of an end-to-end tracking which directly maps from raw sensor data to a fully unoccluded occupancy grid map is introduced by Ondr\'{u}\v{s}ka et~al. in \cite{DBLP:journals/corr/OndruskaP16}, using a synthetic dataset containing circular objects and only linear motion.
This work is extended in \cite{DBLP:journals/corr/OndruskaDWP16} with the use of real-world data that covers a busy urban intersection similar to our scenario with cars, cyclists, and pedestrians.
The data is represented as occupancy grid maps with a size of $100\times100$, containing two binary channels to encode visibility and occupancy, in $20\times20\,\mathrm{cm}^2$ grids.
One should note that our proposed network is trained with a $480\times480$ grid.
The results for our presented approach are compared in Section \ref{sec:Eval} to the Deep Tracking approach in \cite{doi:10.1177/0278364917710543} using real-world data of an urban intersection, recorded from a stationary sensor platform.
\section{System Overview}\label{sec:network}
\subsection{Input Data}
Grid maps represent the $360\degree$ vehicle environment as a spatial discretization in a bird's eye view, where every cell contains information about the space located at its position.
The dynamic occupancy grid maps (DOGMas) \cite{DBLP:journals/corr/NussRTYKMGD16} used as input data in this work employ a Dempster-Shafer \cite{Dempster2008} representation of the occupancy state and a particle filter approach for capturing motion of objects.
The illustration of a DOGMa in Fig. \ref{fig:datastructure2} uses grayscale to encode the occupancy probability $P_O$, where dark pixels refer to occupied cells, white to free space and gray to unobserved area. 
The orientation of dynamic objects, according to their velocity estimate, is illustrated by colors corresponding to the colored circle.
The ego vehicle is the black rectangle in the center.
The DOGMa data is provided in $\mathds{R}^{W \times H \times \Omega}$, where $W$ and $H$ are the spatial width and height. 
The channels $\Omega = \{M_O, M_F, v_E, v_N, \sigma_{VE}^2,\sigma_{VN}^2, \sigma_{VE,VN}^2\}$ contain the mass for occupied $M_O \in [0,1] $ and the mass for free $M_F \in [0,1]$. 
Furthermore, the velocities to the east $v_E$ and north $v_N$ and their variances and covariance are stored.
The occupancy probability $P_O$ of a grid cell is defined as $P_O = M_O + 0.5(1-M_O-M_F)$.
We omit the variances and covariance and use only the channels $M_O, M_F, v_E, v_N$ of the DOGMas as input data.
\subsection{Grid Predictor Model Architecture}
\begin{figure*}[thpb]
	\vspace{1mm}
	\centering
	\includegraphics[width=0.9\textwidth]{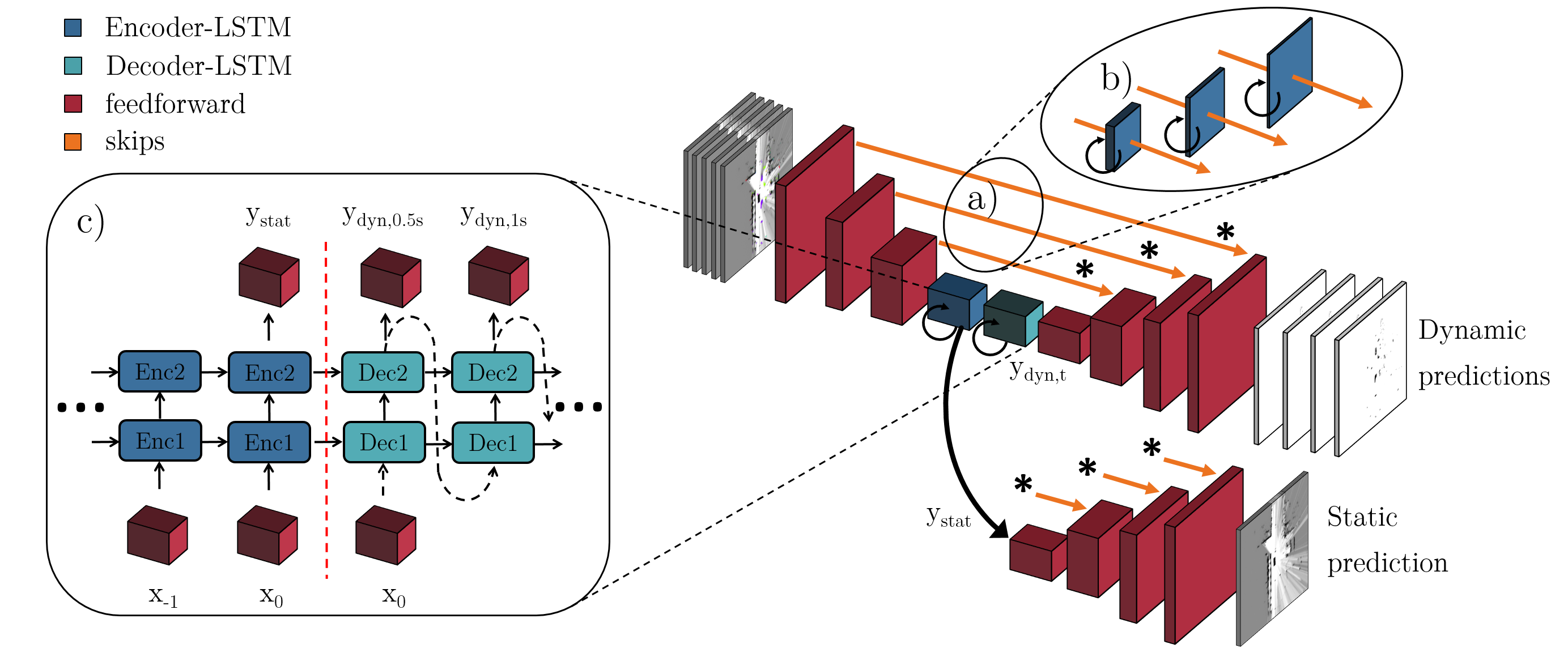}
	\caption{Overview of the developed network architecture. The main parts are the down- and upscaling structure consisting of feedforward neural networks (red) and the recurrent Encoder-Decoder model, outlined in c). In the Grid Predictor Model, the feedforward skip connections are depicted in a), the extended version with recurrent skip connections is outlined in b).}
	\label{fig:network_overview}
\end{figure*}
The developed network architecture is a composition of feedforward neural network parts and recurrent network layers as shown in Fig.~\ref{fig:network_overview}. 
The feedforward network parts, depicted in red are inspired by \cite{DBLP:journals/corr/NohHH15}, where the convolutional layers of the VGG 16\hbox{-}layer net in \cite{DBLP:journals/corr/SimonyanZ14a} are followed by a mirrored deconvolution network.
The use of convolutional neural networks on DOGMas undermines the assumption of independent cells in the particle filter approach and allows the model to use context information for capturing the scenario.
Several convolutional layers reduce the spatial size of the input grid map, provided in $\mathds{R}^{480 \times 480 \times 4}$, to a tensor with the size $\mathds{R}^{18 \times 18 \times 128}$ in three steps.
This tensor is fed in the LSTM\hbox{-}layer, where the predictions are made with the same spatial sizes as the input tensors.
The upscaling of this predictions is performed using learnable transposed convolutions and adding skip connections, as depicted in Fig. \ref{fig:network_overview} a).
The skip connections provide lower feature maps as additional inputs to the upscaling layers as proposed in \cite{DBLP:journals/corr/LongSD14} and \cite{DBLP:journals/corr/RonnebergerFB15} to get dense predictions.
The upscaling structure is built with two separate paths by using transposed convolutional layers with own parameters for the static and dynamic prediction.
For upscaling a single feedforward part is sequentially used for each prediction step.
So, the spatial transformations are the same for different time steps and the prediction task is only executed by the recurrent Encoder-Decoder model, located in the middle of the architecture, with the downscaled feature representation.

The recurrent network part is inspired by the Encoder-Decoder framework introduced by Srivastava et~al. in \cite{DBLP:journals/corr/SrivastavaMS15} based on the general sequence-to-sequence framework in \cite{DBLP:journals/corr/SutskeverVL14}. 
Here, it is used to train the Encoder\hbox{-}LSTM and the Decoder\hbox{-}LSTM with sequences of different length and time scales.
The architecture of the two\hbox{-}layer Encoder-Decoder is depicted in Fig. \ref{fig:network_overview} c) with the unrolled structure during training.
The task of the Encoder\hbox{-}LSTM is to observe the input sequence and save information of several time steps in its internal states, which can be seen as sequential filtering of the scene.
In addition, the output of the Encoder\hbox{-}LSTM is used as static prediction $\boldsymbol{y\textsubscript{stat}}$. 
Each time step, the Decoder\hbox{-}LSTM will be initialized with the current internal states of the Encoder\hbox{-}LSTM and produces a sequence of future predictions. 
In this work, the Decoder\hbox{-}LSTM is applied four times to produce the predictions for $t_0 + 0.5$\,s, $t_0 + 1$\,s, $t_0 + 1.5$\,s and $t_0 + 2$\,s, where $t_0$ indicates the current time.

Both, the Encoder\hbox{-}LSTM and the Decoder\hbox{-}LSTM consist of a two\hbox{-}layer ConvLSTM~\cite{DBLP:journals/corr/ShiCWYWW15} with kernels of the size $5 \times 5$ and internal states $\boldsymbol{h}$, $\boldsymbol{c}$ both in $\mathds{R}^{18 \times 18 \times 128}$.  
Here, the ConvLSTMs exploit spatio-temporal information, but the spatial size of all internal states $\boldsymbol{h}$, $\boldsymbol{c}$ and the outputs $\boldsymbol{y\textsubscript{stat}}$, $\boldsymbol{y\textsubscript{dyn}}$ remain the same. 
The use of kernels instead of a fully connected architecture takes into account that the same motion patterns could be detected on different places and that the movements of the objects are spatially limited.
The developed Grid Predictor Model (GPM) contains 14.65 million parameters and achieves an inference time of 63\,ms on a Nvidia GeForce GTX 1080 Ti, which is sufficient for real-time application.
\subsection{Recurrent Skip Architecture}
The described Grid Predictor Model tends to particularly use the current input grid map due to the non\hbox{-}recurrent skip connections. 
Especially in the case of large occlusions and the tracking of small objects through occlusion, this leads to a degraded performance.
To further improve the prediction performance in these occlusion scenarios, we refined the architecture by inserting ConvLSTMs in every skip connection as shown in Fig. \ref{fig:network_overview} b).
These newly introduced ConvLSTMs are trained to observe the input sequence and provide high resolution features to the layers in the upscaling structure including occluded objects.
The main part of capturing motion and performing the future prediction are still made by the Encoder-Decoder model in the middle of the architecture.
The ConvLSTMs in the skip connections have a reduced number of channels by choosing a lower number of input kernels than input feature maps.
This design choice is a key technique to train the model with long input sequences in terms of memory constraints.  
The effect of this novel recurrent skip architecture is further examined in Section \ref{sec:Eval}.
\section{Dataset and Training}\label{sec:dataset}
We recorded an urban intersection scenario with pedestrians, cyclists, and cars at two different positions and three different days.
From $4$ sequences, each about $30$ minutes, the first 80\,\% are used as training set, the remaining 20\,\% as test set for evaluation.
A shorter sequence was used for validation.
To speed up training, reduce memory consumption, and keep real-time performance the original $901\times901$ grid maps are cropped to the center, leading to input data provided in $\mathds{R}^{480 \times 480 \times 4}$.
However, the preserved part of the map covers most of the perceived dynamic objects and due to the fully convolutional structure, the proposed network is applicable on data of arbitrary size.
Due to the prediction task the desired output can be observed at a later time.
We use the algorithm of \cite{Stumper18LabelExtraction}, to automatically extract labels of the dynamic objects and the static environment separately.
Besides the ability to distinguish between static and dynamic objects, the extracted dynamic labels are used for a cell wise weighting factor to counteract the high imbalance of background and rare dynamic cells.
\subsection{Spatial Balancing Loss}
Our network consists of two output paths to distinguish between the static environment and dynamic objects.    
So, the overall loss $L\textsubscript{o}$ is a weighted sum of the static loss $L\textsubscript{s}$ and the dynamic loss $L\textsubscript{d}$
\begin{equation}
	L\textsubscript{o} = L\textsubscript{s} + k\textsubscript{o} L\textsubscript{d}
\end{equation}
with the weight factor $k\textsubscript{o}$ to determine the balance between the two loss terms.
The loss of the static environment
\begin{equation}
	L\textsubscript{s} = \frac{1}{W \times H} \sum_{c=1}^{W \times H} |y\textsubscript{s}^*(c)-y\textsubscript{s}(c)|
\end{equation} 
is the mean of the absolute differences of each cell $c$ between the static label $y\textsubscript{s}^*$ and the static prediction $y\textsubscript{s}$ of the model. 
The dynamic loss
\begin{equation}
	L\textsubscript{d} = \frac{1}{n\textsubscript{pred}}\frac{1}{W \times H} \sum_{i=1}^{n\textsubscript{pred}}\sum_{c=1}^{W \times H} \lambda_{c,i} (y\textsubscript{d}^*(c,i)-y\textsubscript{d}(c,i))^2
\end{equation}   
is the mean squared error between the dynamic label $y\textsubscript{d}^*$ and the dynamic prediction $y\textsubscript{d}$ over each cell $c$ and each prediction time $i$ and contains a cell wise weighting factor $\lambda_{c,i}$ as proposed in our previous work \cite{DBLP:journals/corr/HoermannBD17}.
For the prediction task, it is crucial to add the cell wise weighting factor $\lambda_{c,i}$ due to the strong imbalance of background and dynamic cells.
The cell wise weighting factor $\lambda_{c,i}$ is calculated according
\begin{equation}
	\lambda_{c,i} = 1 + k_i\cdot y\textsubscript{d}^*(c,i)
\end{equation}  
for each cell $c$ and each prediction horizon $i$.
The multiplication with the dynamic label $y\textsubscript{d}^*(c,i)$ leads to a weighting $\lambda_{c,i} = 1$ for static cells and $\lambda_{c,i} = 1 + k_i$ for dynamic cells.
According to the static/dynamic ratio, we chose $k_i=40$.
\subsection{Data Structure and Training}
In this work the data is stored as continuous sequence of input grid maps at $10$\,Hz, and the matching static and dynamic labels at each time step, as depicted in Fig. \ref{fig:datastructure2}.
So, at a random time step $t_0$ the static label shows the occupancy probabilities of the static environment, and the dynamic label contains solely the dynamic objects of the current input grid map.
During training, the network is fed with a sequence of input grid maps of length $n\textsubscript{in}$.
It outputs the static prediction and $n\textsubscript{pred}$ dynamic predictions with the time span $\Delta t\textsubscript{pred}$ between. 
We found that Truncated Backpropagation \cite{Williams90anefficient}, where a sequence is continuously passed through at training, easily leads to overfitting for large ($30$\,min) sequences.
We assume that this is because the network parameters are continually adjusted to the current scene.
Therefore, we implemented a training process with random sequence start time to achieve better generalization.
In every training iteration a random time step $t_0$ in a random sequence is picked, then the input grid maps and the corresponding labels are selected according the chosen training configuration $n\textsubscript{in}$, $n\textsubscript{pred}$ and $\Delta t\textsubscript{pred}$ as shown in Fig.~\ref{fig:datastructure2}.
The above described data structure is memory efficient and allows varying configurations of $n\textsubscript{in}$, $n\textsubscript{pred}$, and $\Delta t\textsubscript{pred}$. 
In this work we choose $n\textsubscript{pred} = 4$ for the number of predictions and $\Delta t\textsubscript{pred}=0.5$\,s for the time span between them. 
This leads to a maximum prediction horizon of 2\,s.
The length of the input sequence is set to $n\textsubscript{in}=5$ ($0.5$\,s) for most experiments, which allows short training duration.
To train the model for tracking objects through occluded area, we choose $n\textsubscript{in}=20$ ($2.0$\,s), so the model is able to capture long-term dependencies. 
In addition to the random training process, we apply dropout on the non\hbox{-}recurrent connections in the LSTMs as proposed in \cite{DBLP:journals/corr/ZarembaSV14} to avoid overfitting.
For the optimization of the parameters the ADAM \cite{DBLP:journals/corr/KingmaB14} solver with the exponential decay rates $\beta_1 =  0.9$ and $\beta_2 = 0.999$ is used. 
The learning rate is set to 0.0001 and the training process is stopped after one epoch.
\section{Evaluation}\label{sec:Eval}
\begin{figure}
	\vspace{0.5mm}
	\centering
	\newlength\fheight
	\newlength\fwidth
	\hspace{-7mm}
	\setlength\fheight{5.4cm}
	\setlength\fwidth{7cm}
	\input{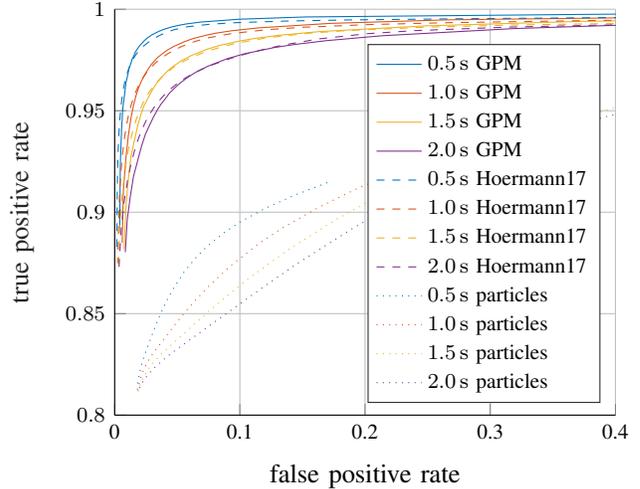}
	\caption[ROC curves of Grid Predictor Model and previous work]{ROC curves of the Grid Predictor Model on the test set and the previous work in \cite{DBLP:journals/corr/HoermannBD17}}
	\label{fig:ROC_comp_h}
\end{figure}
First, the overall prediction performance on the test set is measured with a receiver operating characteristic (ROC) curve and compared to the previous work in \cite{DBLP:journals/corr/HoermannBD17}, then the results are evaluated regarding special characteristics, i.e. the ability to model interactions of objects, produce multimodal predictions and predict occluded objects. 
For the ROC curve, each cell in the predictions and the ground truth has to be classified as occupied or free by applying a certain threshold $\gamma$.
For evaluation ground truth maps are generated, containing the occupancy probabilities for each cell, calculated with the masses for occupied and free in the DOGMa.  
In these ground truth maps all cells with a value above a threshold of $P_O > 0.55$ are classified as occupied, cells with $P_O < 0.45$ as free space. 
So, the unobserved area with values around 0.5 are not used to evaluate the prediction performance.  
To classify occupied cells, for the static prediction a fixed threshold $P_O > 0.5$, for the dynamic prediction a variable threshold $\gamma\textsubscript{dyn} \in (0,1)$ is used. 
The ROC curve is constructed by calculating the true positive rate and the false positive rate for several thresholds $\gamma\textsubscript{dyn}$. 
The leftmost point in the curve is calculated using only the static prediction. 
By reducing the variable threshold $\gamma\textsubscript{dyn}$, more predicted values are interpreted as occupied cells, resulting in an increasing true positive rate, but also a higher rate of false positives.

We compare our recurrent network to our previous work and a Monte Carlo approach in Fig. \ref{fig:ROC_comp_h}.
The figure shows ROC curves for different prediction horizons, where the presented approach (annotated \emph{GPM}) is drawn with solid lines, our previous approach (annotated \emph{Hoermann17}) is drawn with dashed lines, and the Monte Carlo approach (annotated \emph{particles}) with dotted lines.
To get comparable results, we simply cropped the results of the previous work to $480\times480$.
This leads to an advantage on edges, because of the knowledge about the area outside the cropped grid. 
However, the GPM achieves a slightly better performance in the relevant operating points with a false positive rate lower than $0.1$. 
In addition, the Grid Predictor Model contains only 14.65 million parameters, whereas the feedforward neural network in \cite{DBLP:journals/corr/HoermannBD17} has over 220 million parameters.

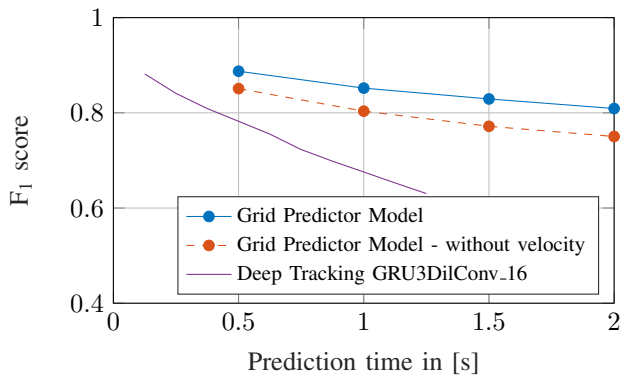
\begin{figure}[thpb]
	\centering
	\setlength\fheight{3.8cm}
	\setlength\fwidth{7cm}
	\hspace{-7mm}
	% This file was created by matlab2tikz.
%
%The latest updates can be retrieved from
%  http://www.mathworks.com/matlabcentral/fileexchange/22022-matlab2tikz-matlab2tikz
%where you can also make suggestions and rate matlab2tikz.
%
\definecolor{mycolor1}{rgb}{0.00000,0.44700,0.74100}%
\definecolor{mycolor2}{rgb}{0.85000,0.32500,0.09800}%
\definecolor{mycolor3}{rgb}{0.92900,0.69400,0.12500}%
\definecolor{mycolor4}{rgb}{0.49400,0.18400,0.55600}%
\begin{tikzpicture}
\begin{axis}[%
width=0.951\fwidth,
height=\fheight,
at={(0\fwidth,0\fheight)},
scale only axis,
xmin=0,
xmax=2,
xlabel style={font=\color{white!15!black}},
xlabel={Prediction time in [s]},
ymin=0.4,
ymax=1,
ylabel style={font=\color{white!15!black}},
ylabel={$\text{F}_\text{1}\text{ score}$},
axis background/.style={fill=white},
xmajorgrids,
ymajorgrids,
legend pos=south east,
legend style={legend cell align=left, align=left, draw=white!15!black},
legend style={font=\footnotesize}
]

\addplot [color=mycolor1, mark=*, mark options={solid, mycolor1}]
  table[row sep=crcr]{%
0.5	0.887350988223234\\
1	0.851898172114534\\
1.5	0.829132801498422\\
2	0.809026932880971\\
};
\addlegendentry{Grid Predictor Model} % wie thesis alles > 0.55

%\addplot [color=mycolor1, mark=*, mark options={solid, mycolor1}]
%table[row sep=crcr]{%
%	0.5	0.8901\\
%	1	0.8561\\
%	1.5	0.8345\\
%	2	0.8150\\
%};
%\addlegendentry{Grid Predictor Model} % mit free <= 0.45 und static >= 0.55

\addplot [color=mycolor2, dashed, mark=*, mark options={solid, mycolor2}]
table[row sep=crcr]{%
0.5	0.8509\\
1	0.8035\\
1.5	0.7717\\
2	0.7503\\
};
\addlegendentry{Grid Predictor Model - without velocity} % wie thesis alles > 0.55

\addplot [color=mycolor4, mark=none, mark options={solid, mycolor3}]
  table[row sep=crcr]{%
0.125  0.8818	\\
0.249  0.8411	\\
0.373  0.8094	\\
0.5    0.7822	\\
0.624  0.7555	\\
0.748  0.7233	\\
0.875  0.6984	\\
1      0.6757	\\
1.123  0.6531	\\
1.25   0.6304	\\
};
\addlegendentry{Deep Tracking GRU3DilConv\_16}

%\addplot [color=mycolor2, mark=*, mark options={solid, mycolor2}]
%table[row sep=crcr]{%
%	0.5	0.889538607284932\\
%	1	0.853032059866749\\
%	1.5	0.829454262148839\\
%	2	0.803726884697224\\
%	2.5	0.772680066270777\\
%	3	0.734447020624061\\
%};
%\addlegendentry{Network}
%
%\addplot [color=mycolor3, dashed, mark=diamond, mark options={solid, mycolor3}]
%table[row sep=crcr]{%
%	0.5	0.793231083900046\\
%	1	0.788376828500561\\
%	1.5	0.784225278204617\\
%	2	0.780955379403015\\
%	2.5	0.777746008169574\\
%	3	0.77469664145208\\
%};
%\addlegendentry{Monte Carlo}

\end{axis}
\end{tikzpicture}%
	\caption[F1-scores]{F1\hbox{-}scores of the Grid Predictor Model, with and without the use of the velocity channels as input, and the Deep Tracking approach in \cite{doi:10.1177/0278364917710543}. }
	\label{fig:F1_scores}
\end{figure}
In the Deep Tracking approach of Dequaire et al., summarized in \cite{doi:10.1177/0278364917710543}, recurrent neural networks are used to predict a fully unoccluded occupancy grid from raw laser data. 
The results of the proposed work are compared to results of the Deep Tracking approach, applied on real-world data with a stationary platform, using the F\textsubscript{1} score. 
Here, we used a fixed threshold of $P_O > 0.55$ to classify cells as occupied in the ground truth map, the static and the dynamic prediction. 
The threshold value is chosen to be slightly above 0.5 to sum up the unobserved gray areas and the free space to the negative class.
The F\textsubscript{1} scores of the best model in \cite{doi:10.1177/0278364917710543} and the Grid Predictor Model are depicted in Fig. \ref{fig:F1_scores} as solid lines. 
It can be seen, that this work outperforms the Deep Tracking approach in a long-term prediction task, but it has to be taken into account that the used data and the tasks are different.
The main goal in the Deep Tracking approach is to track objects through full occlusion and provide an unoccluded grid map as output for each time step, whereas in this work, the focus lies on the long-term prediction.
In addition, the filtered dynamic occupancy grid map, used as input, provides velocity estimates of each cell and therefore simplifies the task of predicting the movements of dynamic objects.

In a further experiment the Grid Predictor Model is trained only using the first two channels $c =\{M_O, M_F\}$ of the DOGMa.
In this setting the network has to detect the motion of objects solely by observing the temporal correlation of the occupied cells.
The dashed red line in Fig. \ref{fig:F1_scores} shows the performance of this experiment. 
As expected, without using the velocity channels the performance decreases.
\subsection{Interaction between Objects}
\begin{figure}[thpb]
		\centering
		\def\svgwidth{0.9\columnwidth}		
		%% Creator: Inkscape inkscape 0.92.3, www.inkscape.org
%% PDF/EPS/PS + LaTeX output extension by Johan Engelen, 2010
%% Accompanies image file 'interaction_input.pdf' (pdf, eps, ps)
%%
%% To include the image in your LaTeX document, write
%%   \input{<filename>.pdf_tex}
%%  instead of
%%   \includegraphics{<filename>.pdf}
%% To scale the image, write
%%   \def\svgwidth{<desired width>}
%%   \input{<filename>.pdf_tex}
%%  instead of
%%   \includegraphics[width=<desired width>]{<filename>.pdf}
%%
%% Images with a different path to the parent latex file can
%% be accessed with the `import' package (which may need to be
%% installed) using
%%   \usepackage{import}
%% in the preamble, and then including the image with
%%   \import{<path to file>}{<filename>.pdf_tex}
%% Alternatively, one can specify
%%   \graphicspath{{<path to file>/}}
%% 
%% For more information, please see info/svg-inkscape on CTAN:
%%   http://tug.ctan.org/tex-archive/info/svg-inkscape
%%
\begingroup%
  \makeatletter%
  \providecommand\color[2][]{%
    \errmessage{(Inkscape) Color is used for the text in Inkscape, but the package 'color.sty' is not loaded}%
    \renewcommand\color[2][]{}%
  }%
  \providecommand\transparent[1]{%
    \errmessage{(Inkscape) Transparency is used (non-zero) for the text in Inkscape, but the package 'transparent.sty' is not loaded}%
    \renewcommand\transparent[1]{}%
  }%
  \providecommand\rotatebox[2]{#2}%
  \newcommand*\fsize{\dimexpr\f@size pt\relax}%
  \newcommand*\lineheight[1]{\fontsize{\fsize}{#1\fsize}\selectfont}%
  \ifx\svgwidth\undefined%
    \setlength{\unitlength}{304.5bp}%
    \ifx\svgscale\undefined%
      \relax%
    \else%
      \setlength{\unitlength}{\unitlength * \real{\svgscale}}%
    \fi%
  \else%
    \setlength{\unitlength}{\svgwidth}%
  \fi%
  \global\let\svgwidth\undefined%
  \global\let\svgscale\undefined%
  \makeatother%
  \begin{picture}(1,0.49261084)%
    \lineheight{1}%
    \setlength\tabcolsep{0pt}%
    \put(0,0){\includegraphics[width=\unitlength,page=1]{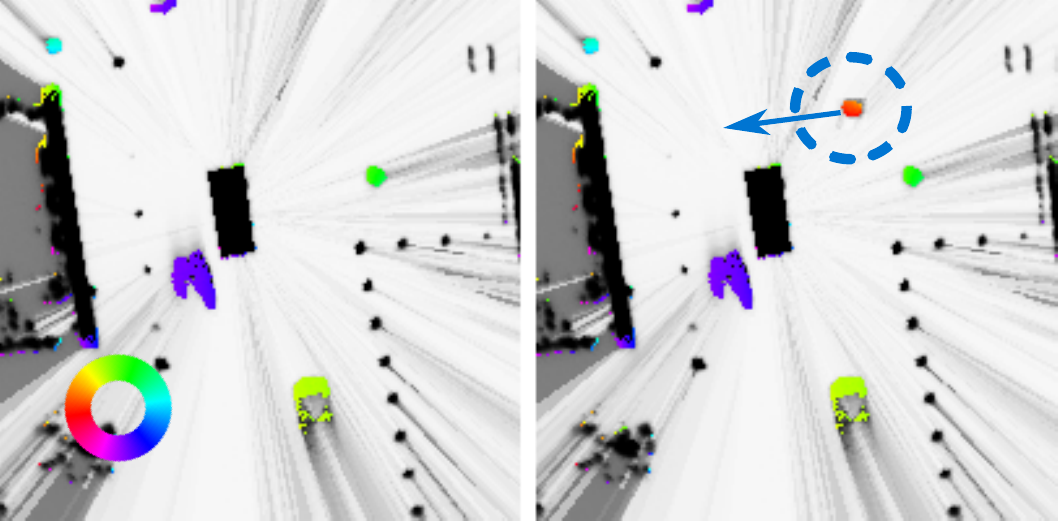}}%
    \put(0.11399788,0.01){\color[rgb]{0,0,0}\makebox(0,0)[t]{\lineheight{1.25}\smash{\begin{tabular}[t]{c}\bfseries orientation\end{tabular}}}}%
  \end{picture}%
\endgroup%

		\\
		\vspace{0.1cm}
		\centering
		\def\svgwidth{0.9\columnwidth}	
		%% Creator: Inkscape inkscape 0.92.3, www.inkscape.org
%% PDF/EPS/PS + LaTeX output extension by Johan Engelen, 2010
%% Accompanies image file 'interaction_pred_15_2.pdf' (pdf, eps, ps)
%%
%% To include the image in your LaTeX document, write
%%   \input{<filename>.pdf_tex}
%%  instead of
%%   \includegraphics{<filename>.pdf}
%% To scale the image, write
%%   \def\svgwidth{<desired width>}
%%   \input{<filename>.pdf_tex}
%%  instead of
%%   \includegraphics[width=<desired width>]{<filename>.pdf}
%%
%% Images with a different path to the parent latex file can
%% be accessed with the `import' package (which may need to be
%% installed) using
%%   \usepackage{import}
%% in the preamble, and then including the image with
%%   \import{<path to file>}{<filename>.pdf_tex}
%% Alternatively, one can specify
%%   \graphicspath{{<path to file>/}}
%% 
%% For more information, please see info/svg-inkscape on CTAN:
%%   http://tug.ctan.org/tex-archive/info/svg-inkscape
%%
\begingroup%
  \makeatletter%
  \providecommand\color[2][]{%
    \errmessage{(Inkscape) Color is used for the text in Inkscape, but the package 'color.sty' is not loaded}%
    \renewcommand\color[2][]{}%
  }%
  \providecommand\transparent[1]{%
    \errmessage{(Inkscape) Transparency is used (non-zero) for the text in Inkscape, but the package 'transparent.sty' is not loaded}%
    \renewcommand\transparent[1]{}%
  }%
  \providecommand\rotatebox[2]{#2}%
  \newcommand*\fsize{\dimexpr\f@size pt\relax}%
  \newcommand*\lineheight[1]{\fontsize{\fsize}{#1\fsize}\selectfont}%
  \ifx\svgwidth\undefined%
    \setlength{\unitlength}{304.5bp}%
    \ifx\svgscale\undefined%
      \relax%
    \else%
      \setlength{\unitlength}{\unitlength * \real{\svgscale}}%
    \fi%
  \else%
    \setlength{\unitlength}{\svgwidth}%
  \fi%
  \global\let\svgwidth\undefined%
  \global\let\svgscale\undefined%
  \makeatother%
  \begin{picture}(1,0.49261084)%
    \lineheight{1}%
    \setlength\tabcolsep{0pt}%
    \put(0,0){\includegraphics[width=\unitlength,page=1]{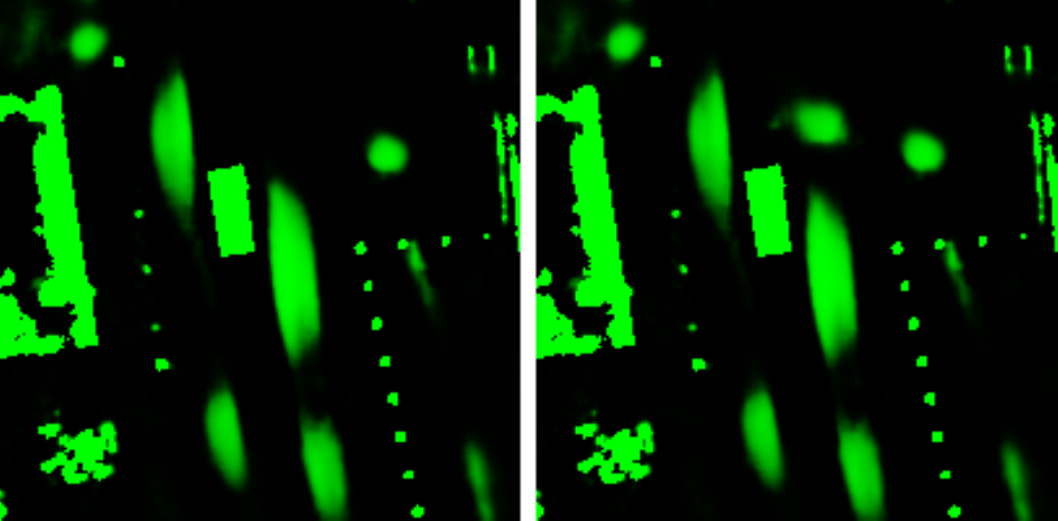}}%
    \put(0.05042535,0.0101304){\color[rgb]{1,1,1}\makebox(0,0)[t]{\lineheight{1.25}\smash{\begin{tabular}[t]{c}\bfseries1.5\,s\end{tabular}}}}%
    \put(0.55781457,0.0101304){\color[rgb]{1,1,1}\makebox(0,0)[t]{\lineheight{1.25}\smash{\begin{tabular}[t]{c}\bfseries1.5\,s\end{tabular}}}}%
    \put(0,0){\includegraphics[width=\unitlength,page=2]{interaction_pred_15_2.pdf}}%
  \end{picture}%
\endgroup%

	\caption[Illustration of interaction between objects]{Illustration of the interaction between objects.
		A pedestrian was artificially placed into a traffic scene to examine prediction adaption.
		 The original (left) and manipulated (right) input grid maps are shown in the top row, the predictions of the network for these inputs below in green. 
		 The artificially added pedestrian, marked with a blue circle, has the intention to cross the road.
		 At $1.5$\,s, the original prediction reaches the red reference line, while the prediction of the manipulated scene does not.
	 }
	\label{fig:interaction}
\end{figure}
The DOGMa that is used as input data contains occupancy and velocity information, that results from a particle filter estimating the occupancy and velocity distribution for each cell.
In the particle filter approach each grid cell is updated independently following the assumption, that there are no interactions between cells. 
The usage of convolutional neural networks undermines this assumption by exploiting spatial context. 
Thus, the developed Grid Predictor Model has the ability to model interactions of objects. 
This behavior is examined by artificially adding a pedestrian to the input grid map, marked with a blue circle in the top right image in Fig. \ref{fig:interaction}.
The artificially added pedestrian has the intention to cross the road, recognizable by the velocity estimation that is directed to the left, according the red color. 
The reaction of the trained network is directly observable by comparing the predictions based on the original and the manipulated input scene.
In Fig. \ref{fig:interaction} the output of the network for the prediction time horizon of 1.5\,s with the original input is shown in the left column, with the manipulated input in the right column.
With the manipulated input, the network takes into account the crossing of the pedestrian and therefore a slowing down of the approaching vehicle is predicted.
\subsection{Multimodal Prediction}
\begin{figure*}[thpb]
	\vspace{1mm}
	\centering
	\begin{subfigure}{0.19\textwidth}
		\def\svgwidth{\columnwidth}		
		%% Creator: Inkscape inkscape 0.92.3, www.inkscape.org
%% PDF/EPS/PS + LaTeX output extension by Johan Engelen, 2010
%% Accompanies image file 'multi_in.pdf' (pdf, eps, ps)
%%
%% To include the image in your LaTeX document, write
%%   \input{<filename>.pdf_tex}
%%  instead of
%%   \includegraphics{<filename>.pdf}
%% To scale the image, write
%%   \def\svgwidth{<desired width>}
%%   \input{<filename>.pdf_tex}
%%  instead of
%%   \includegraphics[width=<desired width>]{<filename>.pdf}
%%
%% Images with a different path to the parent latex file can
%% be accessed with the `import' package (which may need to be
%% installed) using
%%   \usepackage{import}
%% in the preamble, and then including the image with
%%   \import{<path to file>}{<filename>.pdf_tex}
%% Alternatively, one can specify
%%   \graphicspath{{<path to file>/}}
%% 
%% For more information, please see info/svg-inkscape on CTAN:
%%   http://tug.ctan.org/tex-archive/info/svg-inkscape
%%
\begingroup%
  \makeatletter%
  \providecommand\color[2][]{%
    \errmessage{(Inkscape) Color is used for the text in Inkscape, but the package 'color.sty' is not loaded}%
    \renewcommand\color[2][]{}%
  }%
  \providecommand\transparent[1]{%
    \errmessage{(Inkscape) Transparency is used (non-zero) for the text in Inkscape, but the package 'transparent.sty' is not loaded}%
    \renewcommand\transparent[1]{}%
  }%
  \providecommand\rotatebox[2]{#2}%
  \newcommand*\fsize{\dimexpr\f@size pt\relax}%
  \newcommand*\lineheight[1]{\fontsize{\fsize}{#1\fsize}\selectfont}%
  \ifx\svgwidth\undefined%
    \setlength{\unitlength}{191.25bp}%
    \ifx\svgscale\undefined%
      \relax%
    \else%
      \setlength{\unitlength}{\unitlength * \real{\svgscale}}%
    \fi%
  \else%
    \setlength{\unitlength}{\svgwidth}%
  \fi%
  \global\let\svgwidth\undefined%
  \global\let\svgscale\undefined%
  \makeatother%
  \begin{picture}(1,1)%
    \lineheight{1}%
    \setlength\tabcolsep{0pt}%
    \put(0,0){\includegraphics[width=\unitlength,page=1]{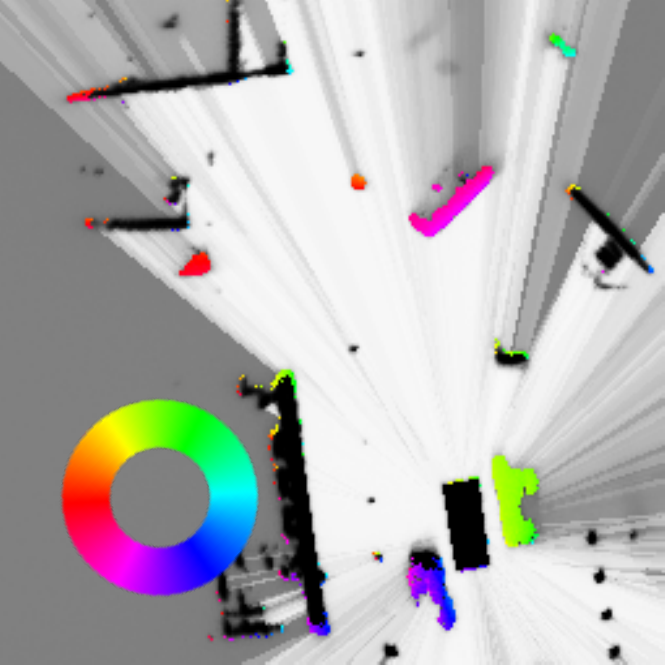}}%
    \put(0.01392936,0.01637638){\color[rgb]{1,1,1}\makebox(0,0)[lt]{\lineheight{1.25}\smash{\begin{tabular}[t]{l}\bfseries orientation\end{tabular}}}}%
    \put(0,0){\includegraphics[width=\unitlength,page=2]{multi_in.pdf}}%
  \end{picture}%
\endgroup%

	\end{subfigure}
	\begin{subfigure}{0.19\textwidth}
		\def\svgwidth{\columnwidth}		
		%% Creator: Inkscape inkscape 0.92.3, www.inkscape.org
%% PDF/EPS/PS + LaTeX output extension by Johan Engelen, 2010
%% Accompanies image file 'multi_05.pdf' (pdf, eps, ps)
%%
%% To include the image in your LaTeX document, write
%%   \input{<filename>.pdf_tex}
%%  instead of
%%   \includegraphics{<filename>.pdf}
%% To scale the image, write
%%   \def\svgwidth{<desired width>}
%%   \input{<filename>.pdf_tex}
%%  instead of
%%   \includegraphics[width=<desired width>]{<filename>.pdf}
%%
%% Images with a different path to the parent latex file can
%% be accessed with the `import' package (which may need to be
%% installed) using
%%   \usepackage{import}
%% in the preamble, and then including the image with
%%   \import{<path to file>}{<filename>.pdf_tex}
%% Alternatively, one can specify
%%   \graphicspath{{<path to file>/}}
%% 
%% For more information, please see info/svg-inkscape on CTAN:
%%   http://tug.ctan.org/tex-archive/info/svg-inkscape
%%
\begingroup%
  \makeatletter%
  \providecommand\color[2][]{%
    \errmessage{(Inkscape) Color is used for the text in Inkscape, but the package 'color.sty' is not loaded}%
    \renewcommand\color[2][]{}%
  }%
  \providecommand\transparent[1]{%
    \errmessage{(Inkscape) Transparency is used (non-zero) for the text in Inkscape, but the package 'transparent.sty' is not loaded}%
    \renewcommand\transparent[1]{}%
  }%
  \providecommand\rotatebox[2]{#2}%
  \newcommand*\fsize{\dimexpr\f@size pt\relax}%
  \newcommand*\lineheight[1]{\fontsize{\fsize}{#1\fsize}\selectfont}%
  \ifx\svgwidth\undefined%
    \setlength{\unitlength}{191.25bp}%
    \ifx\svgscale\undefined%
      \relax%
    \else%
      \setlength{\unitlength}{\unitlength * \real{\svgscale}}%
    \fi%
  \else%
    \setlength{\unitlength}{\svgwidth}%
  \fi%
  \global\let\svgwidth\undefined%
  \global\let\svgscale\undefined%
  \makeatother%
  \begin{picture}(1,1)%
    \lineheight{1}%
    \setlength\tabcolsep{0pt}%
    \put(0,0){\includegraphics[width=\unitlength,page=1]{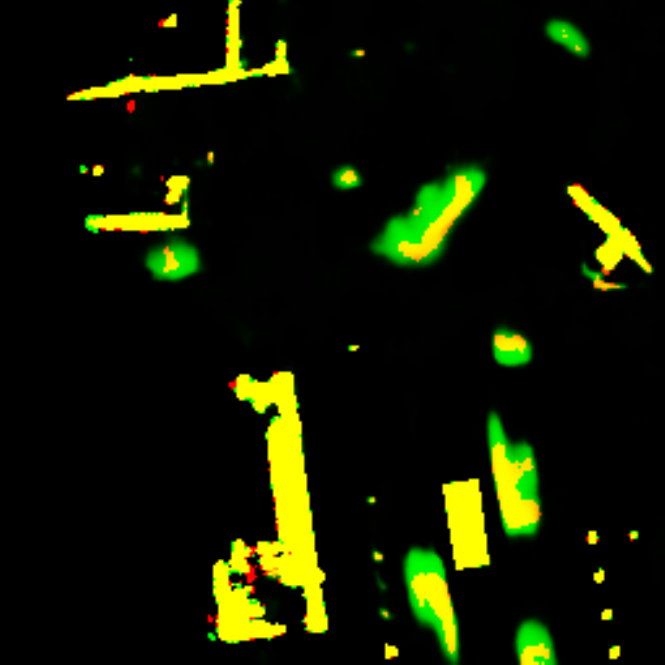}}%
    \put(0.02,0.02){\color[rgb]{1,1,1}\makebox(0,0)[lt]{\lineheight{1.25}\smash{\begin{tabular}[t]{l}\bfseries0.5\,s\end{tabular}}}}%
    \put(0,0){\includegraphics[width=\unitlength,page=2]{multi_05.pdf}}%
  \end{picture}%
\endgroup%

	\end{subfigure}
	\begin{subfigure}{0.19\textwidth}
		\def\svgwidth{\columnwidth}		
		%% Creator: Inkscape inkscape 0.92.3, www.inkscape.org
%% PDF/EPS/PS + LaTeX output extension by Johan Engelen, 2010
%% Accompanies image file 'multi_10.pdf' (pdf, eps, ps)
%%
%% To include the image in your LaTeX document, write
%%   \input{<filename>.pdf_tex}
%%  instead of
%%   \includegraphics{<filename>.pdf}
%% To scale the image, write
%%   \def\svgwidth{<desired width>}
%%   \input{<filename>.pdf_tex}
%%  instead of
%%   \includegraphics[width=<desired width>]{<filename>.pdf}
%%
%% Images with a different path to the parent latex file can
%% be accessed with the `import' package (which may need to be
%% installed) using
%%   \usepackage{import}
%% in the preamble, and then including the image with
%%   \import{<path to file>}{<filename>.pdf_tex}
%% Alternatively, one can specify
%%   \graphicspath{{<path to file>/}}
%% 
%% For more information, please see info/svg-inkscape on CTAN:
%%   http://tug.ctan.org/tex-archive/info/svg-inkscape
%%
\begingroup%
  \makeatletter%
  \providecommand\color[2][]{%
    \errmessage{(Inkscape) Color is used for the text in Inkscape, but the package 'color.sty' is not loaded}%
    \renewcommand\color[2][]{}%
  }%
  \providecommand\transparent[1]{%
    \errmessage{(Inkscape) Transparency is used (non-zero) for the text in Inkscape, but the package 'transparent.sty' is not loaded}%
    \renewcommand\transparent[1]{}%
  }%
  \providecommand\rotatebox[2]{#2}%
  \newcommand*\fsize{\dimexpr\f@size pt\relax}%
  \newcommand*\lineheight[1]{\fontsize{\fsize}{#1\fsize}\selectfont}%
  \ifx\svgwidth\undefined%
    \setlength{\unitlength}{191.25bp}%
    \ifx\svgscale\undefined%
      \relax%
    \else%
      \setlength{\unitlength}{\unitlength * \real{\svgscale}}%
    \fi%
  \else%
    \setlength{\unitlength}{\svgwidth}%
  \fi%
  \global\let\svgwidth\undefined%
  \global\let\svgscale\undefined%
  \makeatother%
  \begin{picture}(1,1)%
    \lineheight{1}%
    \setlength\tabcolsep{0pt}%
    \put(0,0){\includegraphics[width=\unitlength,page=1]{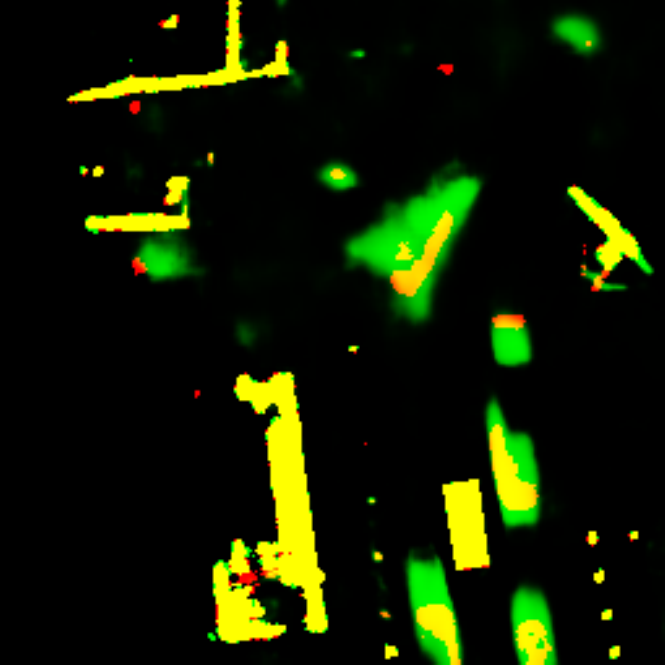}}%
    \put(0.02,0.02){\color[rgb]{1,1,1}\makebox(0,0)[lt]{\lineheight{1.25}\smash{\begin{tabular}[t]{l}\bfseries1\,s\end{tabular}}}}%
    \put(0,0){\includegraphics[width=\unitlength,page=2]{multi_10.pdf}}%
  \end{picture}%
\endgroup%

	\end{subfigure}
	\begin{subfigure}{0.19\textwidth}
		\def\svgwidth{\columnwidth}		
		%% Creator: Inkscape inkscape 0.92.3, www.inkscape.org
%% PDF/EPS/PS + LaTeX output extension by Johan Engelen, 2010
%% Accompanies image file 'multi_15.pdf' (pdf, eps, ps)
%%
%% To include the image in your LaTeX document, write
%%   \input{<filename>.pdf_tex}
%%  instead of
%%   \includegraphics{<filename>.pdf}
%% To scale the image, write
%%   \def\svgwidth{<desired width>}
%%   \input{<filename>.pdf_tex}
%%  instead of
%%   \includegraphics[width=<desired width>]{<filename>.pdf}
%%
%% Images with a different path to the parent latex file can
%% be accessed with the `import' package (which may need to be
%% installed) using
%%   \usepackage{import}
%% in the preamble, and then including the image with
%%   \import{<path to file>}{<filename>.pdf_tex}
%% Alternatively, one can specify
%%   \graphicspath{{<path to file>/}}
%% 
%% For more information, please see info/svg-inkscape on CTAN:
%%   http://tug.ctan.org/tex-archive/info/svg-inkscape
%%
\begingroup%
  \makeatletter%
  \providecommand\color[2][]{%
    \errmessage{(Inkscape) Color is used for the text in Inkscape, but the package 'color.sty' is not loaded}%
    \renewcommand\color[2][]{}%
  }%
  \providecommand\transparent[1]{%
    \errmessage{(Inkscape) Transparency is used (non-zero) for the text in Inkscape, but the package 'transparent.sty' is not loaded}%
    \renewcommand\transparent[1]{}%
  }%
  \providecommand\rotatebox[2]{#2}%
  \newcommand*\fsize{\dimexpr\f@size pt\relax}%
  \newcommand*\lineheight[1]{\fontsize{\fsize}{#1\fsize}\selectfont}%
  \ifx\svgwidth\undefined%
    \setlength{\unitlength}{191.25bp}%
    \ifx\svgscale\undefined%
      \relax%
    \else%
      \setlength{\unitlength}{\unitlength * \real{\svgscale}}%
    \fi%
  \else%
    \setlength{\unitlength}{\svgwidth}%
  \fi%
  \global\let\svgwidth\undefined%
  \global\let\svgscale\undefined%
  \makeatother%
  \begin{picture}(1,1)%
    \lineheight{1}%
    \setlength\tabcolsep{0pt}%
    \put(0,0){\includegraphics[width=\unitlength,page=1]{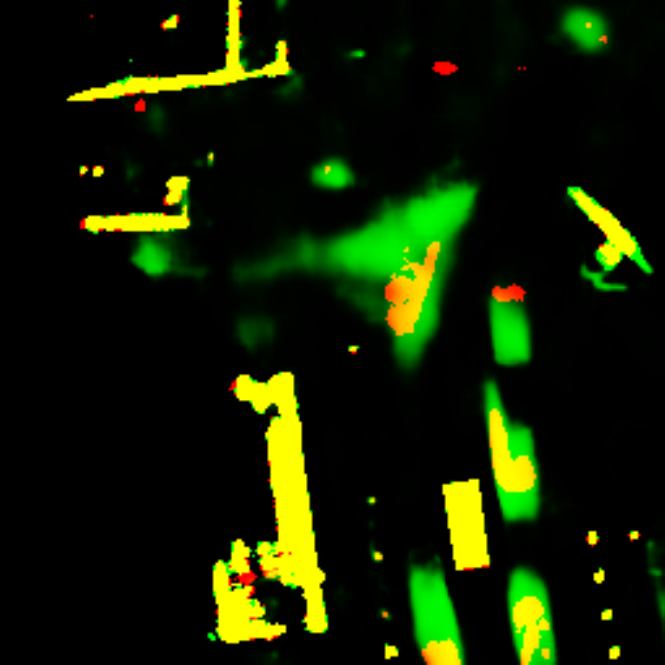}}%
    \put(0.02,0.02){\color[rgb]{1,1,1}\makebox(0,0)[lt]{\lineheight{1.25}\smash{\begin{tabular}[t]{l}\bfseries1.5\,s\end{tabular}}}}%
    \put(0,0){\includegraphics[width=\unitlength,page=2]{multi_15.pdf}}%
  \end{picture}%
\endgroup%

	\end{subfigure}
	\begin{subfigure}{0.19\textwidth}
		\def\svgwidth{\columnwidth}		
		%% Creator: Inkscape inkscape 0.92.3, www.inkscape.org
%% PDF/EPS/PS + LaTeX output extension by Johan Engelen, 2010
%% Accompanies image file 'multi_20.pdf' (pdf, eps, ps)
%%
%% To include the image in your LaTeX document, write
%%   \input{<filename>.pdf_tex}
%%  instead of
%%   \includegraphics{<filename>.pdf}
%% To scale the image, write
%%   \def\svgwidth{<desired width>}
%%   \input{<filename>.pdf_tex}
%%  instead of
%%   \includegraphics[width=<desired width>]{<filename>.pdf}
%%
%% Images with a different path to the parent latex file can
%% be accessed with the `import' package (which may need to be
%% installed) using
%%   \usepackage{import}
%% in the preamble, and then including the image with
%%   \import{<path to file>}{<filename>.pdf_tex}
%% Alternatively, one can specify
%%   \graphicspath{{<path to file>/}}
%% 
%% For more information, please see info/svg-inkscape on CTAN:
%%   http://tug.ctan.org/tex-archive/info/svg-inkscape
%%
\begingroup%
  \makeatletter%
  \providecommand\color[2][]{%
    \errmessage{(Inkscape) Color is used for the text in Inkscape, but the package 'color.sty' is not loaded}%
    \renewcommand\color[2][]{}%
  }%
  \providecommand\transparent[1]{%
    \errmessage{(Inkscape) Transparency is used (non-zero) for the text in Inkscape, but the package 'transparent.sty' is not loaded}%
    \renewcommand\transparent[1]{}%
  }%
  \providecommand\rotatebox[2]{#2}%
  \newcommand*\fsize{\dimexpr\f@size pt\relax}%
  \newcommand*\lineheight[1]{\fontsize{\fsize}{#1\fsize}\selectfont}%
  \ifx\svgwidth\undefined%
    \setlength{\unitlength}{191.25bp}%
    \ifx\svgscale\undefined%
      \relax%
    \else%
      \setlength{\unitlength}{\unitlength * \real{\svgscale}}%
    \fi%
  \else%
    \setlength{\unitlength}{\svgwidth}%
  \fi%
  \global\let\svgwidth\undefined%
  \global\let\svgscale\undefined%
  \makeatother%
  \begin{picture}(1,1)%
    \lineheight{1}%
    \setlength\tabcolsep{0pt}%
    \put(0,0){\includegraphics[width=\unitlength,page=1]{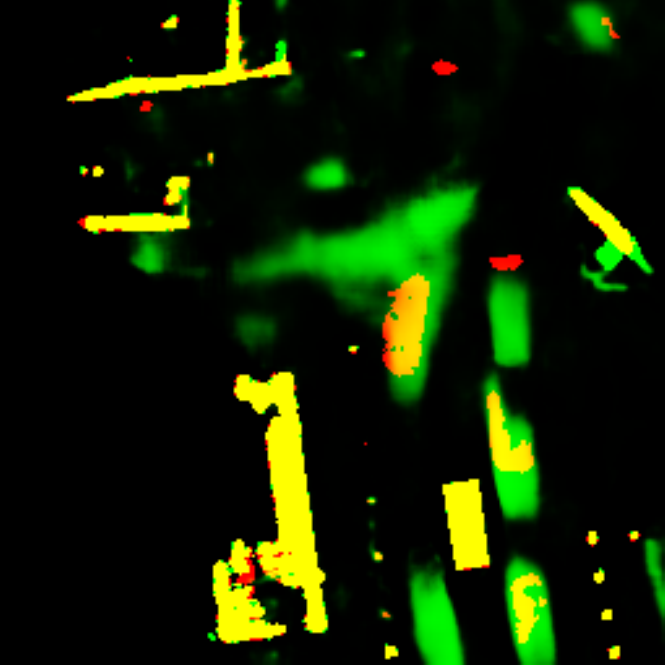}}%
    \put(0.02,0.02){\color[rgb]{1,1,1}\makebox(0,0)[lt]{\lineheight{1.25}\smash{\begin{tabular}[t]{l}\bfseries2\,s\end{tabular}}}}%
    \put(0,0){\includegraphics[width=\unitlength,page=2]{multi_20.pdf}}%
  \end{picture}%
\endgroup%

	\end{subfigure}
	\caption[Illustration of multimodal prediction]{Illustration of the capability to produce multimodal predictions. 
		A turning vehicle is marked with a dashed blue circle and appears purple in the input DOGMa (left).
		The four images on the right illustrate the prediction results. 
		In these RGB-images, the prediction is stored in the green, the ground truth in the red channel. This leads to the following interpretation of the colors: green: false positive, yellow: true positive, red: false negative. 
		At input time the vehicle could either go left or follow the road down. 
		After 1\,s the network produces a multimodal prediction, i.e. two possible paths of the vehicle.}
	\label{fig:multimodal}
\end{figure*}
In crossing scenarios in the city center, vehicles commonly have the options to turn left or right or drive straight ahead.
Thus, the long-term prediction of vehicles on crossroads is a complex task, which is not deterministically solvable.
The simple forward propagation of occupied cells according the velocity estimation in the DOGMa only covers one possible option, which sometimes is obviously wrong.
In contrast, the Grid Predictor Model has the ability to produce multiple hypotheses for the future position of an object.
In the excerpt of the input grid map in Fig. \ref{fig:multimodal} the marked vehicle drives diagonally to the bottom left and has the possibility to turn left or drive straight ahead.
In contrast, the estimated velocity orientation points to the bottom left, so following this estimation with a constant velocity would lead to an average prediction of the two possible modes.
The four right images depict the results for the four prediction time horizons $t_0 + 0.5$\,s, $t_0 + 1$\,s, $t_0 + 1.5$\,s and $t_0 + 2$\,s, where $t_0$ indicates the current time.
The green RGB-channel is used for the outputs of the neural network, the red channel for the ground truth.
So, the resulting grid maps show the true positive as yellow, false positives in green and false negatives in red.
The prediction results in Fig. \ref{fig:multimodal} show that the network has the ability to produce multimodal predictions.
\subsection{Generalization Capability}
\begin{figure*}[thpb]
	\vspace{1mm}
	\centering
	\captionsetup[subfigure]{labelformat=empty}
	%\hspace{0.04\textwidth}
	\begin{subfigure}{0.4\columnwidth}
		\centering
		\def\svgwidth{\columnwidth}		
		%% Creator: Inkscape inkscape 0.92.3, www.inkscape.org
%% PDF/EPS/PS + LaTeX output extension by Johan Engelen, 2010
%% Accompanies image file 'in_train.pdf' (pdf, eps, ps)
%%
%% To include the image in your LaTeX document, write
%%   \input{<filename>.pdf_tex}
%%  instead of
%%   \includegraphics{<filename>.pdf}
%% To scale the image, write
%%   \def\svgwidth{<desired width>}
%%   \input{<filename>.pdf_tex}
%%  instead of
%%   \includegraphics[width=<desired width>]{<filename>.pdf}
%%
%% Images with a different path to the parent latex file can
%% be accessed with the `import' package (which may need to be
%% installed) using
%%   \usepackage{import}
%% in the preamble, and then including the image with
%%   \import{<path to file>}{<filename>.pdf_tex}
%% Alternatively, one can specify
%%   \graphicspath{{<path to file>/}}
%% 
%% For more information, please see info/svg-inkscape on CTAN:
%%   http://tug.ctan.org/tex-archive/info/svg-inkscape
%%
\begingroup%
  \makeatletter%
  \providecommand\color[2][]{%
    \errmessage{(Inkscape) Color is used for the text in Inkscape, but the package 'color.sty' is not loaded}%
    \renewcommand\color[2][]{}%
  }%
  \providecommand\transparent[1]{%
    \errmessage{(Inkscape) Transparency is used (non-zero) for the text in Inkscape, but the package 'transparent.sty' is not loaded}%
    \renewcommand\transparent[1]{}%
  }%
  \providecommand\rotatebox[2]{#2}%
  \newcommand*\fsize{\dimexpr\f@size pt\relax}%
  \newcommand*\lineheight[1]{\fontsize{\fsize}{#1\fsize}\selectfont}%
  \ifx\svgwidth\undefined%
    \setlength{\unitlength}{360bp}%
    \ifx\svgscale\undefined%
      \relax%
    \else%
      \setlength{\unitlength}{\unitlength * \real{\svgscale}}%
    \fi%
  \else%
    \setlength{\unitlength}{\svgwidth}%
  \fi%
  \global\let\svgwidth\undefined%
  \global\let\svgscale\undefined%
  \makeatother%
  \begin{picture}(1,1)%
    \lineheight{1}%
    \setlength\tabcolsep{0pt}%
    \put(0,0){\includegraphics[width=\unitlength,page=1]{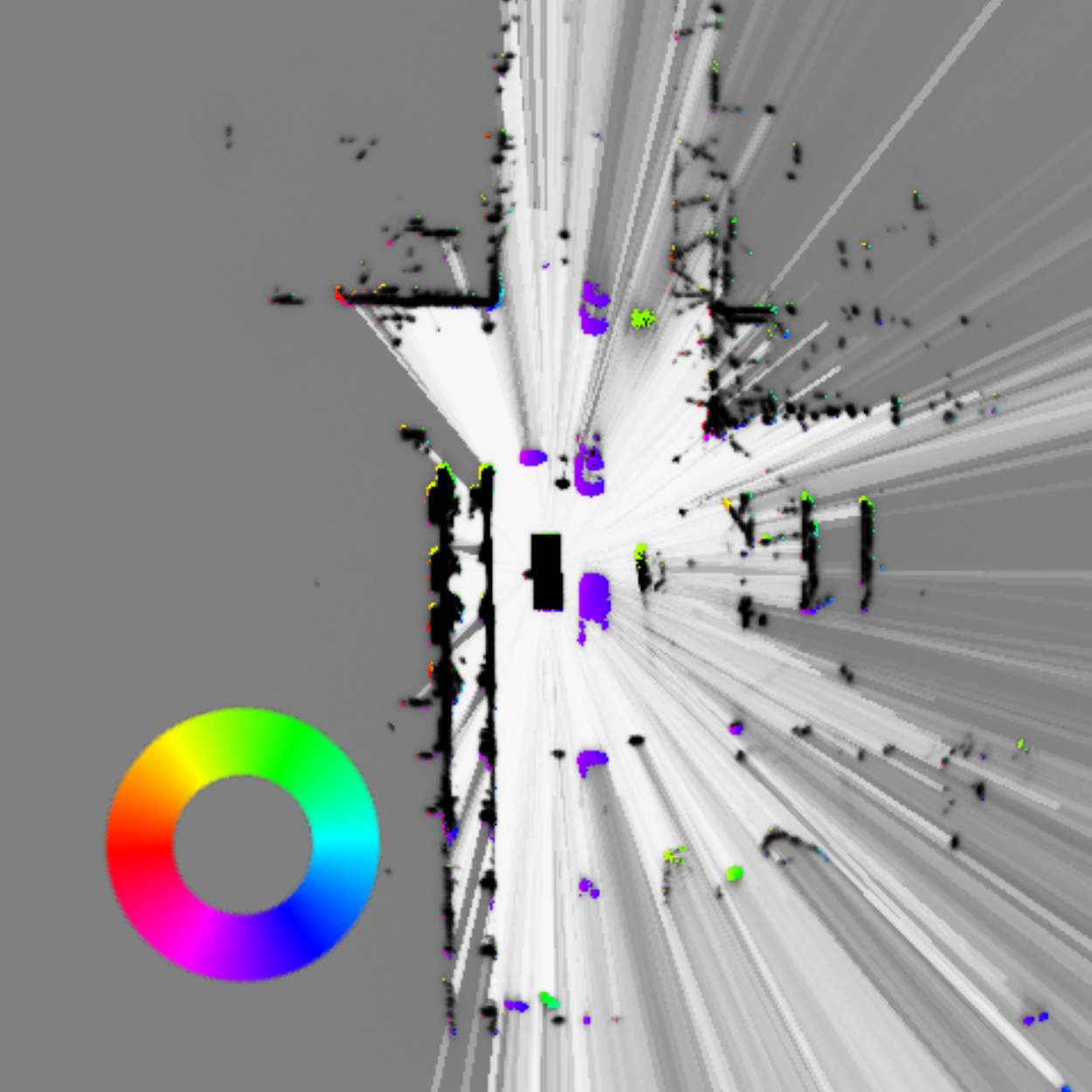}}%
    \put(0.01363925,0.01657639){\color[rgb]{1,1,1}\makebox(0,0)[lt]{\lineheight{1.25}\smash{\begin{tabular}[t]{l}\bfseries orientation\end{tabular}}}}%
  \end{picture}%
\endgroup%

		\caption{training data}
	\end{subfigure}
	\hspace{0.04\textwidth}
	%\quad
	\begin{subfigure}{0.4\columnwidth}
		\centering
		\def\svgwidth{\columnwidth}		
		%% Creator: Inkscape inkscape 0.92.3, www.inkscape.org
%% PDF/EPS/PS + LaTeX output extension by Johan Engelen, 2010
%% Accompanies image file 'in_test.pdf' (pdf, eps, ps)
%%
%% To include the image in your LaTeX document, write
%%   \input{<filename>.pdf_tex}
%%  instead of
%%   \includegraphics{<filename>.pdf}
%% To scale the image, write
%%   \def\svgwidth{<desired width>}
%%   \input{<filename>.pdf_tex}
%%  instead of
%%   \includegraphics[width=<desired width>]{<filename>.pdf}
%%
%% Images with a different path to the parent latex file can
%% be accessed with the `import' package (which may need to be
%% installed) using
%%   \usepackage{import}
%% in the preamble, and then including the image with
%%   \import{<path to file>}{<filename>.pdf_tex}
%% Alternatively, one can specify
%%   \graphicspath{{<path to file>/}}
%% 
%% For more information, please see info/svg-inkscape on CTAN:
%%   http://tug.ctan.org/tex-archive/info/svg-inkscape
%%
\begingroup%
  \makeatletter%
  \providecommand\color[2][]{%
    \errmessage{(Inkscape) Color is used for the text in Inkscape, but the package 'color.sty' is not loaded}%
    \renewcommand\color[2][]{}%
  }%
  \providecommand\transparent[1]{%
    \errmessage{(Inkscape) Transparency is used (non-zero) for the text in Inkscape, but the package 'transparent.sty' is not loaded}%
    \renewcommand\transparent[1]{}%
  }%
  \providecommand\rotatebox[2]{#2}%
  \newcommand*\fsize{\dimexpr\f@size pt\relax}%
  \newcommand*\lineheight[1]{\fontsize{\fsize}{#1\fsize}\selectfont}%
  \ifx\svgwidth\undefined%
    \setlength{\unitlength}{360bp}%
    \ifx\svgscale\undefined%
      \relax%
    \else%
      \setlength{\unitlength}{\unitlength * \real{\svgscale}}%
    \fi%
  \else%
    \setlength{\unitlength}{\svgwidth}%
  \fi%
  \global\let\svgwidth\undefined%
  \global\let\svgscale\undefined%
  \makeatother%
  \begin{picture}(1,1)%
    \lineheight{1}%
    \setlength\tabcolsep{0pt}%
    \put(0,0){\includegraphics[width=\unitlength,page=1]{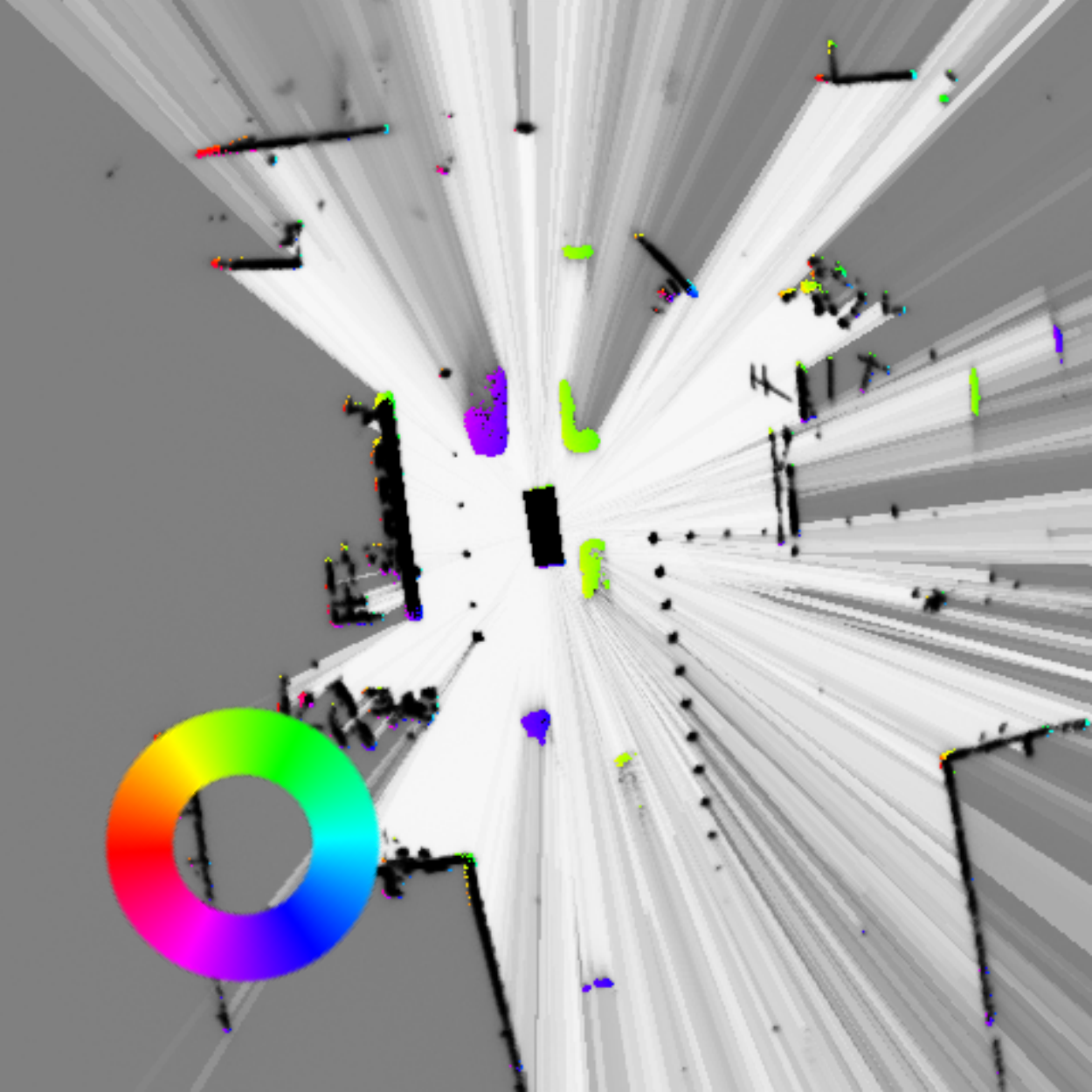}}%
    \put(0.01363926,0.01657627){\color[rgb]{1,1,1}\makebox(0,0)[lt]{\lineheight{1.25}\smash{\begin{tabular}[t]{l}\bfseries orientation\end{tabular}}}}%
  \end{picture}%
\endgroup%

		\caption{test data}
	\end{subfigure}
	\hspace{0.04\textwidth}
	\begin{subfigure}{0.4\columnwidth}
		\centering
		\def\svgwidth{\columnwidth}		
		%% Creator: Inkscape inkscape 0.92.3, www.inkscape.org
%% PDF/EPS/PS + LaTeX output extension by Johan Engelen, 2010
%% Accompanies image file 'pred_05.pdf' (pdf, eps, ps)
%%
%% To include the image in your LaTeX document, write
%%   \input{<filename>.pdf_tex}
%%  instead of
%%   \includegraphics{<filename>.pdf}
%% To scale the image, write
%%   \def\svgwidth{<desired width>}
%%   \input{<filename>.pdf_tex}
%%  instead of
%%   \includegraphics[width=<desired width>]{<filename>.pdf}
%%
%% Images with a different path to the parent latex file can
%% be accessed with the `import' package (which may need to be
%% installed) using
%%   \usepackage{import}
%% in the preamble, and then including the image with
%%   \import{<path to file>}{<filename>.pdf_tex}
%% Alternatively, one can specify
%%   \graphicspath{{<path to file>/}}
%% 
%% For more information, please see info/svg-inkscape on CTAN:
%%   http://tug.ctan.org/tex-archive/info/svg-inkscape
%%
\begingroup%
  \makeatletter%
  \providecommand\color[2][]{%
    \errmessage{(Inkscape) Color is used for the text in Inkscape, but the package 'color.sty' is not loaded}%
    \renewcommand\color[2][]{}%
  }%
  \providecommand\transparent[1]{%
    \errmessage{(Inkscape) Transparency is used (non-zero) for the text in Inkscape, but the package 'transparent.sty' is not loaded}%
    \renewcommand\transparent[1]{}%
  }%
  \providecommand\rotatebox[2]{#2}%
  \newcommand*\fsize{\dimexpr\f@size pt\relax}%
  \newcommand*\lineheight[1]{\fontsize{\fsize}{#1\fsize}\selectfont}%
  \ifx\svgwidth\undefined%
    \setlength{\unitlength}{360bp}%
    \ifx\svgscale\undefined%
      \relax%
    \else%
      \setlength{\unitlength}{\unitlength * \real{\svgscale}}%
    \fi%
  \else%
    \setlength{\unitlength}{\svgwidth}%
  \fi%
  \global\let\svgwidth\undefined%
  \global\let\svgscale\undefined%
  \makeatother%
  \begin{picture}(1,1)%
    \lineheight{1}%
    \setlength\tabcolsep{0pt}%
    \put(0,0){\includegraphics[width=\unitlength,page=1]{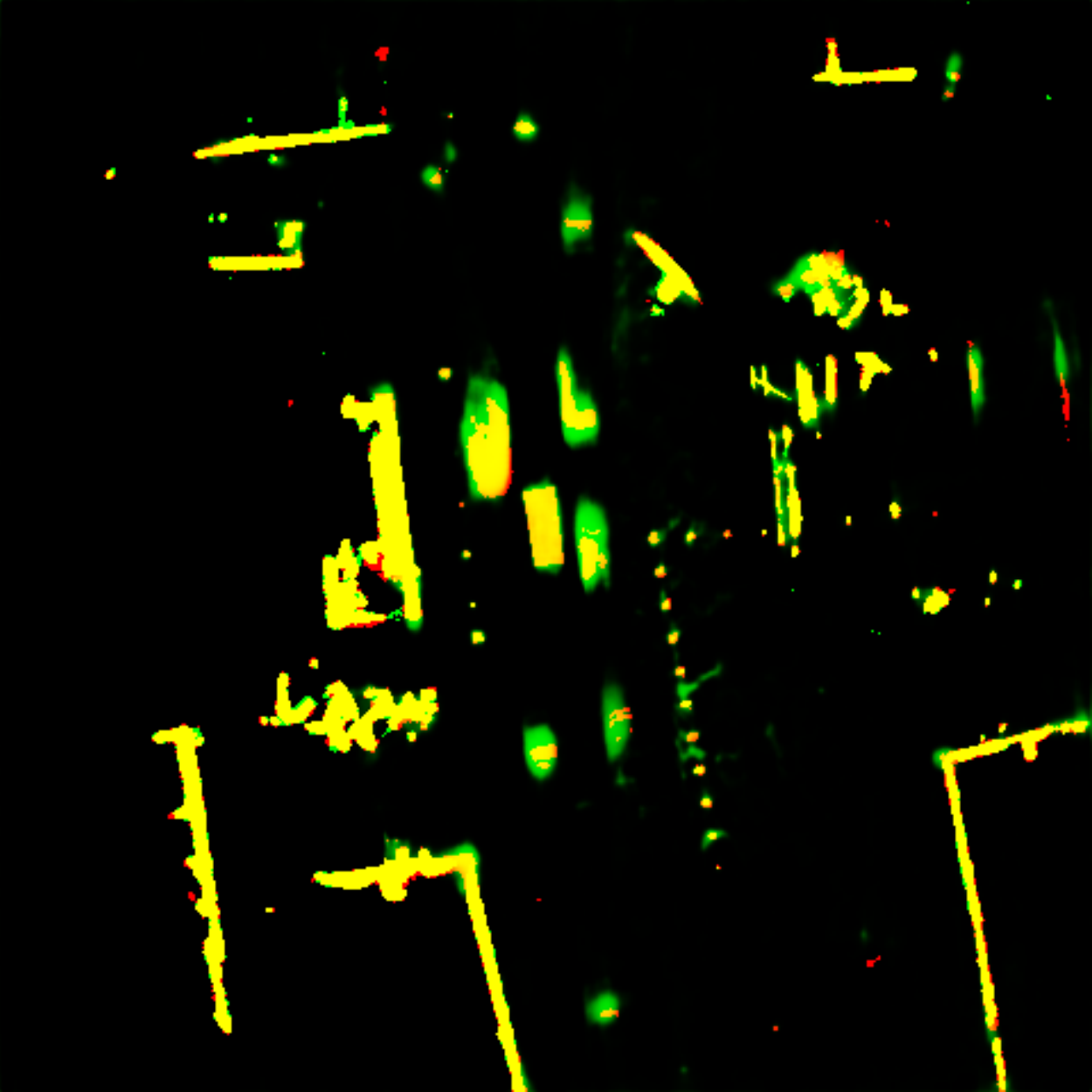}}%
    \put(0.02045573,0.02487951){\color[rgb]{1,1,1}\makebox(0,0)[lt]{\lineheight{1.25}\smash{\begin{tabular}[t]{l}\bfseries0.5\,s\end{tabular}}}}%
  \end{picture}%
\endgroup%

		\caption{}
	\end{subfigure}
	\hspace{0.04\textwidth}
	%\quad
	\begin{subfigure}{0.4\columnwidth}
		\centering
		\def\svgwidth{\columnwidth}		
		%% Creator: Inkscape inkscape 0.92.3, www.inkscape.org
%% PDF/EPS/PS + LaTeX output extension by Johan Engelen, 2010
%% Accompanies image file 'pred_15.pdf' (pdf, eps, ps)
%%
%% To include the image in your LaTeX document, write
%%   \input{<filename>.pdf_tex}
%%  instead of
%%   \includegraphics{<filename>.pdf}
%% To scale the image, write
%%   \def\svgwidth{<desired width>}
%%   \input{<filename>.pdf_tex}
%%  instead of
%%   \includegraphics[width=<desired width>]{<filename>.pdf}
%%
%% Images with a different path to the parent latex file can
%% be accessed with the `import' package (which may need to be
%% installed) using
%%   \usepackage{import}
%% in the preamble, and then including the image with
%%   \import{<path to file>}{<filename>.pdf_tex}
%% Alternatively, one can specify
%%   \graphicspath{{<path to file>/}}
%% 
%% For more information, please see info/svg-inkscape on CTAN:
%%   http://tug.ctan.org/tex-archive/info/svg-inkscape
%%
\begingroup%
  \makeatletter%
  \providecommand\color[2][]{%
    \errmessage{(Inkscape) Color is used for the text in Inkscape, but the package 'color.sty' is not loaded}%
    \renewcommand\color[2][]{}%
  }%
  \providecommand\transparent[1]{%
    \errmessage{(Inkscape) Transparency is used (non-zero) for the text in Inkscape, but the package 'transparent.sty' is not loaded}%
    \renewcommand\transparent[1]{}%
  }%
  \providecommand\rotatebox[2]{#2}%
  \newcommand*\fsize{\dimexpr\f@size pt\relax}%
  \newcommand*\lineheight[1]{\fontsize{\fsize}{#1\fsize}\selectfont}%
  \ifx\svgwidth\undefined%
    \setlength{\unitlength}{360bp}%
    \ifx\svgscale\undefined%
      \relax%
    \else%
      \setlength{\unitlength}{\unitlength * \real{\svgscale}}%
    \fi%
  \else%
    \setlength{\unitlength}{\svgwidth}%
  \fi%
  \global\let\svgwidth\undefined%
  \global\let\svgscale\undefined%
  \makeatother%
  \begin{picture}(1,1)%
    \lineheight{1}%
    \setlength\tabcolsep{0pt}%
    \put(0,0){\includegraphics[width=\unitlength,page=1]{pred_15.pdf}}%
    \put(0.02045573,0.02487957){\color[rgb]{1,1,1}\makebox(0,0)[lt]{\lineheight{1.25}\smash{\begin{tabular}[t]{l}\bfseries 1.5\,s\end{tabular}}}}%
  \end{picture}%
\endgroup%

		\caption{}
	\end{subfigure}
	\caption[Test of the generalization ability]{Illustration of the capability to generalize on different intersections. 
		The two left images show the scenes used for training and evaluation. 
		The GPM is able to produce quite good predictions for the time horizon 0.5\,s and the static environment. At 1.5\,s the model underestimates the movements of the objects on the left side of the ego vehicle.}
	\label{fig:generalization}
\end{figure*}

In this work, we recorded an urban intersection from two different positions and use the last 20\,\% of each sequence as test set for evaluation.
With this experimental setting, it is not possible to make an assumption of the performance on unseen data.
Therefore, we did an experiment, in which one scenario is used for training and the other one for evaluation to test the generalization capability.
The scenario for training is illustrated in the left image in Fig. \ref{fig:generalization}, the scenario for evaluation next to it.
The scenario for evaluation shows a completely different static environment.
Furthermore, the ego vehicle is located between two driving lanes, which are rotated for approximately $20\degree$ compared to the training scenario. 
The two images on the right show the prediction results for the time horizons $t_0 + 0.5$\,s and $t_0 + 1.5$\,s.
The Grid Predictor Model is able to make dense predictions of unseen static environment.
Also the predictions of dynamic objects are mainly correct, but the model has difficulties with the vehicles driving downwards on the left side of the ego vehicle.
In Table \ref{tab:generalization} the F\textsubscript{1} scores of the Grid Predictor Model using all data for training is compared with the results of this experiment, named as one scenario.
As expected, the F\textsubscript{1} scores decrease, especially for longer time horizons, but the performance is still remarkable. 
\begin{table}
	\centering
	\caption{F\textsubscript{1} scores}
	\label{tab:generalization}
	\begin{tabular}{@{}ccccc@{}}
		\toprule
		training data & 0.5\,s & 1\,s & 1.5\,s & 2\,s \\ \midrule
		two scenarios & 88.73\,\% & 85.18\,\% & 82.91\,\% & 80.90\,\% \\
		one scenario  & 85.83\,\% & 80.32\,\% & 76.27\,\% & 72.90\,\% \\
		\bottomrule
	\end{tabular}
\end{table}
\subsection{Occluded Dynamic Objects}
\begin{figure*}[thpb]
	\centering
	\captionsetup[subfigure]{labelformat=empty}
	\begin{subfigure}{0.4\columnwidth}
		\centering
		\def\svgwidth{\columnwidth}		
		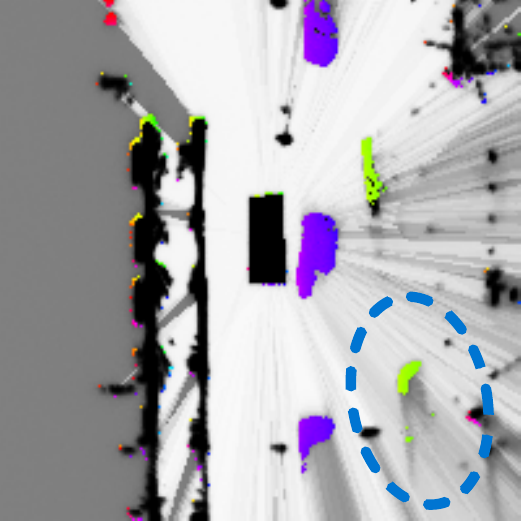
		\caption{t\textsubscript{0} - 0.6\,s}
	\end{subfigure}
	\begin{subfigure}{0.19\columnwidth}
		\centering
		\def\svgwidth{\columnwidth}		
		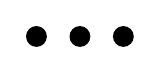
		\caption{}
	\end{subfigure}
	\begin{subfigure}{0.4\columnwidth}
		\centering
		\def\svgwidth{\columnwidth}		
		%% Creator: Inkscape inkscape 0.92.3, www.inkscape.org
%% PDF/EPS/PS + LaTeX output extension by Johan Engelen, 2010
%% Accompanies image file 'input_3578_0.pdf' (pdf, eps, ps)
%%
%% To include the image in your LaTeX document, write
%%   \input{<filename>.pdf_tex}
%%  instead of
%%   \includegraphics{<filename>.pdf}
%% To scale the image, write
%%   \def\svgwidth{<desired width>}
%%   \input{<filename>.pdf_tex}
%%  instead of
%%   \includegraphics[width=<desired width>]{<filename>.pdf}
%%
%% Images with a different path to the parent latex file can
%% be accessed with the `import' package (which may need to be
%% installed) using
%%   \usepackage{import}
%% in the preamble, and then including the image with
%%   \import{<path to file>}{<filename>.pdf_tex}
%% Alternatively, one can specify
%%   \graphicspath{{<path to file>/}}
%% 
%% For more information, please see info/svg-inkscape on CTAN:
%%   http://tug.ctan.org/tex-archive/info/svg-inkscape
%%
\begingroup%
  \makeatletter%
  \providecommand\color[2][]{%
    \errmessage{(Inkscape) Color is used for the text in Inkscape, but the package 'color.sty' is not loaded}%
    \renewcommand\color[2][]{}%
  }%
  \providecommand\transparent[1]{%
    \errmessage{(Inkscape) Transparency is used (non-zero) for the text in Inkscape, but the package 'transparent.sty' is not loaded}%
    \renewcommand\transparent[1]{}%
  }%
  \providecommand\rotatebox[2]{#2}%
  \newcommand*\fsize{\dimexpr\f@size pt\relax}%
  \newcommand*\lineheight[1]{\fontsize{\fsize}{#1\fsize}\selectfont}%
  \ifx\svgwidth\undefined%
    \setlength{\unitlength}{150bp}%
    \ifx\svgscale\undefined%
      \relax%
    \else%
      \setlength{\unitlength}{\unitlength * \real{\svgscale}}%
    \fi%
  \else%
    \setlength{\unitlength}{\svgwidth}%
  \fi%
  \global\let\svgwidth\undefined%
  \global\let\svgscale\undefined%
  \makeatother%
  \begin{picture}(1,1)%
    \lineheight{1}%
    \setlength\tabcolsep{0pt}%
    \put(0,0){\includegraphics[width=\unitlength,page=1]{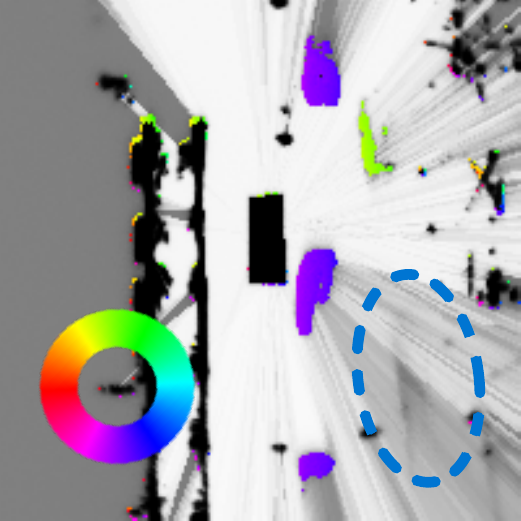}}%
    \put(0.02,0.02){\color[rgb]{1,1,1}\makebox(0,0)[lt]{\lineheight{1.25}\smash{\begin{tabular}[t]{l}\bfseries orientation\end{tabular}}}}%
  \end{picture}%
\endgroup%

		\caption{t\textsubscript{0}}
	\end{subfigure}
	\qquad
	\hspace{1.6cm}
	\begin{subfigure}{0.4\columnwidth}
		\centering
		\def\svgwidth{\columnwidth}		
		%% Creator: Inkscape inkscape 0.92.3, www.inkscape.org
%% PDF/EPS/PS + LaTeX output extension by Johan Engelen, 2010
%% Accompanies image file 'pred_15.pdf' (pdf, eps, ps)
%%
%% To include the image in your LaTeX document, write
%%   \input{<filename>.pdf_tex}
%%  instead of
%%   \includegraphics{<filename>.pdf}
%% To scale the image, write
%%   \def\svgwidth{<desired width>}
%%   \input{<filename>.pdf_tex}
%%  instead of
%%   \includegraphics[width=<desired width>]{<filename>.pdf}
%%
%% Images with a different path to the parent latex file can
%% be accessed with the `import' package (which may need to be
%% installed) using
%%   \usepackage{import}
%% in the preamble, and then including the image with
%%   \import{<path to file>}{<filename>.pdf_tex}
%% Alternatively, one can specify
%%   \graphicspath{{<path to file>/}}
%% 
%% For more information, please see info/svg-inkscape on CTAN:
%%   http://tug.ctan.org/tex-archive/info/svg-inkscape
%%
\begingroup%
  \makeatletter%
  \providecommand\color[2][]{%
    \errmessage{(Inkscape) Color is used for the text in Inkscape, but the package 'color.sty' is not loaded}%
    \renewcommand\color[2][]{}%
  }%
  \providecommand\transparent[1]{%
    \errmessage{(Inkscape) Transparency is used (non-zero) for the text in Inkscape, but the package 'transparent.sty' is not loaded}%
    \renewcommand\transparent[1]{}%
  }%
  \providecommand\rotatebox[2]{#2}%
  \newcommand*\fsize{\dimexpr\f@size pt\relax}%
  \newcommand*\lineheight[1]{\fontsize{\fsize}{#1\fsize}\selectfont}%
  \ifx\svgwidth\undefined%
    \setlength{\unitlength}{150.0000036bp}%
    \ifx\svgscale\undefined%
      \relax%
    \else%
      \setlength{\unitlength}{\unitlength * \real{\svgscale}}%
    \fi%
  \else%
    \setlength{\unitlength}{\svgwidth}%
  \fi%
  \global\let\svgwidth\undefined%
  \global\let\svgscale\undefined%
  \makeatother%
  \begin{picture}(1,1)%
    \lineheight{1}%
    \setlength\tabcolsep{0pt}%
    \put(0,0){\includegraphics[width=\unitlength,page=1]{pred_15.pdf}}%
    \put(0.02,0.02){\color[rgb]{1,1,1}\makebox(0,0)[lt]{\lineheight{1.25}\smash{\begin{tabular}[t]{l}\bfseries1.5\,s\end{tabular}}}}%
  \end{picture}%
\endgroup%

		\caption{Prediction}
	\end{subfigure}
	\hspace{1.7cm}
	\caption[Prediction of occluded vehicle]{Illustration of the capability to make predictions for an occluded vehicle. 
		The input sequence is illustrated with the two images on the left, where the vehicle marked with a blue circle becomes occluded at $t_0$. 
		The Grid Predictor Model is capable to predict the occluded vehicle.}
	\label{fig:occ_vehicle}
\end{figure*}
The input grid maps contain unobserved area indicated with gray color. 
In a scenario with several moving objects, vehicles can be partially or fully occluded as illustrated in Fig. \ref{fig:occ_vehicle}.
The vehicle marked with a blue circle is visible with a velocity estimation at the time $t_0 - 0.6$\,s. 
At the current time $t_0$ there is only a gray silhouette of the vehicle without a velocity estimation.
However, the network developed in this work has the ability to memorize and propagate the object within the internal states of the Encoder\hbox{-}LSTM, even if the subsequent grid maps provide no further information about the now occluded object.
The prediction in Fig. \ref{fig:occ_vehicle} shows that the network is capable to predict the occluded vehicle.

It was found, that it is easier for the model to predict occluded objects, when there is at least a sparse silhouette of it in the current input grid map as shown in Fig. \ref{fig:occ_vehicle} at $t_0$. 
This information of the current input grid map is provided to the upscaling layers by the skip connections and is used for the right localization.
In addition, these skip connections are in particular useful to preserve small objects, such as pedestrians.
Especially to predict small \emph{occluded} dynamic objects, \emph{recurrent} skip connections are used, as evaluated in the following section.
\subsection{Influence of the Recurrent Skip Architecture}
\begin{figure*}[thpb]
	\centering
	\captionsetup[subfigure]{labelformat=empty}
	\begin{subfigure}{0.4\columnwidth}
		\centering
		\def\svgwidth{\columnwidth}		
		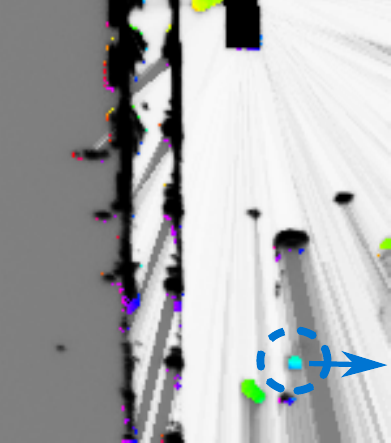
		\caption{t\textsubscript{0} - 0.7\,s}
	\end{subfigure}
	\begin{subfigure}{0.19\columnwidth}
	\centering
	\def\svgwidth{\columnwidth}		
	\import{img/raw/occ_pedestrian/}{points.pdf_tex}
	\caption{}
	\end{subfigure}
	\begin{subfigure}{0.4\columnwidth}
	\centering
	\def\svgwidth{\columnwidth}		
	%% Creator: Inkscape inkscape 0.92.3, www.inkscape.org
%% PDF/EPS/PS + LaTeX output extension by Johan Engelen, 2010
%% Accompanies image file 'occ_ped_in_00.pdf' (pdf, eps, ps)
%%
%% To include the image in your LaTeX document, write
%%   \input{<filename>.pdf_tex}
%%  instead of
%%   \includegraphics{<filename>.pdf}
%% To scale the image, write
%%   \def\svgwidth{<desired width>}
%%   \input{<filename>.pdf_tex}
%%  instead of
%%   \includegraphics[width=<desired width>]{<filename>.pdf}
%%
%% Images with a different path to the parent latex file can
%% be accessed with the `import' package (which may need to be
%% installed) using
%%   \usepackage{import}
%% in the preamble, and then including the image with
%%   \import{<path to file>}{<filename>.pdf_tex}
%% Alternatively, one can specify
%%   \graphicspath{{<path to file>/}}
%% 
%% For more information, please see info/svg-inkscape on CTAN:
%%   http://tug.ctan.org/tex-archive/info/svg-inkscape
%%
\begingroup%
  \makeatletter%
  \providecommand\color[2][]{%
    \errmessage{(Inkscape) Color is used for the text in Inkscape, but the package 'color.sty' is not loaded}%
    \renewcommand\color[2][]{}%
  }%
  \providecommand\transparent[1]{%
    \errmessage{(Inkscape) Transparency is used (non-zero) for the text in Inkscape, but the package 'transparent.sty' is not loaded}%
    \renewcommand\transparent[1]{}%
  }%
  \providecommand\rotatebox[2]{#2}%
  \newcommand*\fsize{\dimexpr\f@size pt\relax}%
  \newcommand*\lineheight[1]{\fontsize{\fsize}{#1\fsize}\selectfont}%
  \ifx\svgwidth\undefined%
    \setlength{\unitlength}{112.5bp}%
    \ifx\svgscale\undefined%
      \relax%
    \else%
      \setlength{\unitlength}{\unitlength * \real{\svgscale}}%
    \fi%
  \else%
    \setlength{\unitlength}{\svgwidth}%
  \fi%
  \global\let\svgwidth\undefined%
  \global\let\svgscale\undefined%
  \makeatother%
  \begin{picture}(1,1.13333337)%
    \lineheight{1}%
    \setlength\tabcolsep{0pt}%
    \put(0,0){\includegraphics[width=\unitlength,page=1]{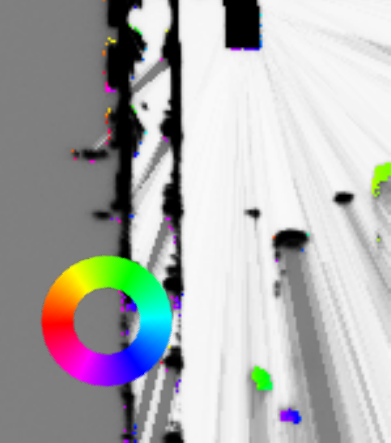}}%
    \put(0.02,0.02){\color[rgb]{1,1,1}\makebox(0,0)[lt]{\lineheight{1.25}\smash{\begin{tabular}[t]{l}\bfseries orientation\end{tabular}}}}%
    \put(0,0){\includegraphics[width=\unitlength,page=2]{occ_ped_in_00.pdf}}%
  \end{picture}%
\endgroup%

	\caption{t\textsubscript{0}}
	\end{subfigure}
	\qquad
	\begin{subfigure}{0.4\columnwidth}
		\centering
		\def\svgwidth{\columnwidth}		
		%% Creator: Inkscape inkscape 0.92.3, www.inkscape.org
%% PDF/EPS/PS + LaTeX output extension by Johan Engelen, 2010
%% Accompanies image file 'pred_15.pdf' (pdf, eps, ps)
%%
%% To include the image in your LaTeX document, write
%%   \input{<filename>.pdf_tex}
%%  instead of
%%   \includegraphics{<filename>.pdf}
%% To scale the image, write
%%   \def\svgwidth{<desired width>}
%%   \input{<filename>.pdf_tex}
%%  instead of
%%   \includegraphics[width=<desired width>]{<filename>.pdf}
%%
%% Images with a different path to the parent latex file can
%% be accessed with the `import' package (which may need to be
%% installed) using
%%   \usepackage{import}
%% in the preamble, and then including the image with
%%   \import{<path to file>}{<filename>.pdf_tex}
%% Alternatively, one can specify
%%   \graphicspath{{<path to file>/}}
%% 
%% For more information, please see info/svg-inkscape on CTAN:
%%   http://tug.ctan.org/tex-archive/info/svg-inkscape
%%
\begingroup%
  \makeatletter%
  \providecommand\color[2][]{%
    \errmessage{(Inkscape) Color is used for the text in Inkscape, but the package 'color.sty' is not loaded}%
    \renewcommand\color[2][]{}%
  }%
  \providecommand\transparent[1]{%
    \errmessage{(Inkscape) Transparency is used (non-zero) for the text in Inkscape, but the package 'transparent.sty' is not loaded}%
    \renewcommand\transparent[1]{}%
  }%
  \providecommand\rotatebox[2]{#2}%
  \newcommand*\fsize{\dimexpr\f@size pt\relax}%
  \newcommand*\lineheight[1]{\fontsize{\fsize}{#1\fsize}\selectfont}%
  \ifx\svgwidth\undefined%
    \setlength{\unitlength}{112.5bp}%
    \ifx\svgscale\undefined%
      \relax%
    \else%
      \setlength{\unitlength}{\unitlength * \real{\svgscale}}%
    \fi%
  \else%
    \setlength{\unitlength}{\svgwidth}%
  \fi%
  \global\let\svgwidth\undefined%
  \global\let\svgscale\undefined%
  \makeatother%
  \begin{picture}(1,1.13333337)%
    \lineheight{1}%
    \setlength\tabcolsep{0pt}%
    \put(0,0){\includegraphics[width=\unitlength,page=1]{pred_15.pdf}}%
    \put(0.02,0.02){\color[rgb]{1,1,1}\makebox(0,0)[lt]{\lineheight{1.25}\smash{\begin{tabular}[t]{l}\bfseries1.5 s\end{tabular}}}}%
    \put(0,0){\includegraphics[width=\unitlength,page=2]{pred_15.pdf}}%
  \end{picture}%
\endgroup%

		\caption{Feedforward skips}
	\end{subfigure}
	\begin{subfigure}{0.4\columnwidth}	
		\centering
		\def\svgwidth{\columnwidth}		
		%% Creator: Inkscape inkscape 0.92.3, www.inkscape.org
%% PDF/EPS/PS + LaTeX output extension by Johan Engelen, 2010
%% Accompanies image file 'pred_15_rec.pdf' (pdf, eps, ps)
%%
%% To include the image in your LaTeX document, write
%%   \input{<filename>.pdf_tex}
%%  instead of
%%   \includegraphics{<filename>.pdf}
%% To scale the image, write
%%   \def\svgwidth{<desired width>}
%%   \input{<filename>.pdf_tex}
%%  instead of
%%   \includegraphics[width=<desired width>]{<filename>.pdf}
%%
%% Images with a different path to the parent latex file can
%% be accessed with the `import' package (which may need to be
%% installed) using
%%   \usepackage{import}
%% in the preamble, and then including the image with
%%   \import{<path to file>}{<filename>.pdf_tex}
%% Alternatively, one can specify
%%   \graphicspath{{<path to file>/}}
%% 
%% For more information, please see info/svg-inkscape on CTAN:
%%   http://tug.ctan.org/tex-archive/info/svg-inkscape
%%
\begingroup%
  \makeatletter%
  \providecommand\color[2][]{%
    \errmessage{(Inkscape) Color is used for the text in Inkscape, but the package 'color.sty' is not loaded}%
    \renewcommand\color[2][]{}%
  }%
  \providecommand\transparent[1]{%
    \errmessage{(Inkscape) Transparency is used (non-zero) for the text in Inkscape, but the package 'transparent.sty' is not loaded}%
    \renewcommand\transparent[1]{}%
  }%
  \providecommand\rotatebox[2]{#2}%
  \newcommand*\fsize{\dimexpr\f@size pt\relax}%
  \newcommand*\lineheight[1]{\fontsize{\fsize}{#1\fsize}\selectfont}%
  \ifx\svgwidth\undefined%
    \setlength{\unitlength}{112.5bp}%
    \ifx\svgscale\undefined%
      \relax%
    \else%
      \setlength{\unitlength}{\unitlength * \real{\svgscale}}%
    \fi%
  \else%
    \setlength{\unitlength}{\svgwidth}%
  \fi%
  \global\let\svgwidth\undefined%
  \global\let\svgscale\undefined%
  \makeatother%
  \begin{picture}(1,1.13333337)%
    \lineheight{1}%
    \setlength\tabcolsep{0pt}%
    \put(0,0){\includegraphics[width=\unitlength,page=1]{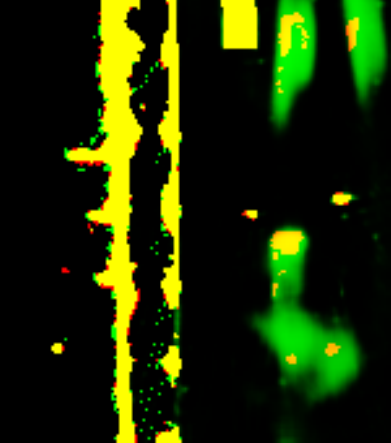}}%
    \put(0.02,0.02){\color[rgb]{1,1,1}\makebox(0,0)[lt]{\lineheight{1.25}\smash{\begin{tabular}[t]{l}\bfseries1.5 s\end{tabular}}}}%
    \put(0,0){\includegraphics[width=\unitlength,page=2]{pred_15_rec.pdf}}%
  \end{picture}%
\endgroup%

		\caption{Recurrent skips}
	\end{subfigure}
	\caption[Prediction of occluded pedestrian]{Illustration of the capability to make predictions for an occluded pedestrian. In the input sequence (two images on the left), the light blue pedestrian is completely occluded at $t_0$. 
		Only the Grid Predictor Model with recurrent skip connections (far right image) is capable to make a prediction for the occluded pedestrian.}
	\label{fig:occ_pedestrian}
\end{figure*}
To tackle the problem of predicting small occluded objects, the Grid Predictor Model is extended with recurrent skip connections as described in Section \ref{sec:network}.
A sequence of input grid maps, in which a pedestrian enters unobserved area is shown in Fig. \ref{fig:occ_pedestrian}.
In the left input grid map at the time $t_0-0.7$\,s the pedestrian, marked with a blue circle, goes to the right and enters the occluded space.
In the current input grid map this pedestrian is fully occluded.
In Fig. \ref{fig:occ_pedestrian} the prediction results of the baseline (third image) are directly compared to the Grid Predictor Model with recurrent skips (fourth image) for the prediction time horizon of $t_0 + 1.5$\,s. 
It is shown, that the Grid Predictor Model with feedforward skips is able to make predictions for the non\hbox{-}occluded objects, but fails in the prediction of the occluded pedestrian (blue circle). 
In contrast, the Grid Predictor Model with recurrent skips is capable to predict the occluded pedestrian.
Thus, the inserted ConvLSTMs in every skip connection enable the ability to track even small occluded objects.

\begin{figure*}[thpb]
	\vspace{1mm}
	\centering
	\captionsetup[subfigure]{labelformat=empty}
	\begin{subfigure}{0.4\columnwidth}
		\centering
		\def\svgwidth{\columnwidth}		
		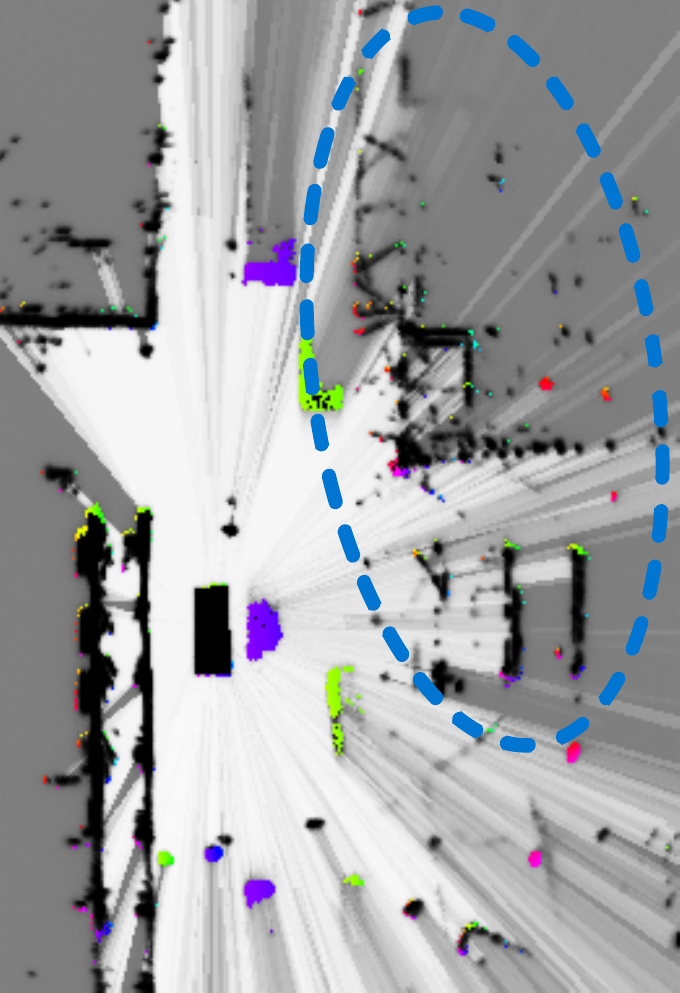
		\caption{t\textsubscript{0} - 5.5\,s}
	\end{subfigure}
	\begin{subfigure}{0.19\columnwidth}
		\centering
		\def\svgwidth{\columnwidth}		
		\import{img/raw/occ_pedestrian/}{points.pdf_tex}
	\end{subfigure}
	\begin{subfigure}{0.4\columnwidth}
		\centering
		\def\svgwidth{\columnwidth}		
		%% Creator: Inkscape inkscape 0.92.3, www.inkscape.org
%% PDF/EPS/PS + LaTeX output extension by Johan Engelen, 2010
%% Accompanies image file 'in_615.pdf' (pdf, eps, ps)
%%
%% To include the image in your LaTeX document, write
%%   \input{<filename>.pdf_tex}
%%  instead of
%%   \includegraphics{<filename>.pdf}
%% To scale the image, write
%%   \def\svgwidth{<desired width>}
%%   \input{<filename>.pdf_tex}
%%  instead of
%%   \includegraphics[width=<desired width>]{<filename>.pdf}
%%
%% Images with a different path to the parent latex file can
%% be accessed with the `import' package (which may need to be
%% installed) using
%%   \usepackage{import}
%% in the preamble, and then including the image with
%%   \import{<path to file>}{<filename>.pdf_tex}
%% Alternatively, one can specify
%%   \graphicspath{{<path to file>/}}
%% 
%% For more information, please see info/svg-inkscape on CTAN:
%%   http://tug.ctan.org/tex-archive/info/svg-inkscape
%%
\begingroup%
  \makeatletter%
  \providecommand\color[2][]{%
    \errmessage{(Inkscape) Color is used for the text in Inkscape, but the package 'color.sty' is not loaded}%
    \renewcommand\color[2][]{}%
  }%
  \providecommand\transparent[1]{%
    \errmessage{(Inkscape) Transparency is used (non-zero) for the text in Inkscape, but the package 'transparent.sty' is not loaded}%
    \renewcommand\transparent[1]{}%
  }%
  \providecommand\rotatebox[2]{#2}%
  \newcommand*\fsize{\dimexpr\f@size pt\relax}%
  \newcommand*\lineheight[1]{\fontsize{\fsize}{#1\fsize}\selectfont}%
  \ifx\svgwidth\undefined%
    \setlength{\unitlength}{195.75bp}%
    \ifx\svgscale\undefined%
      \relax%
    \else%
      \setlength{\unitlength}{\unitlength * \real{\svgscale}}%
    \fi%
  \else%
    \setlength{\unitlength}{\svgwidth}%
  \fi%
  \global\let\svgwidth\undefined%
  \global\let\svgscale\undefined%
  \makeatother%
  \begin{picture}(1,1.45977011)%
    \lineheight{1}%
    \setlength\tabcolsep{0pt}%
    \put(0,0){\includegraphics[width=\unitlength,page=1]{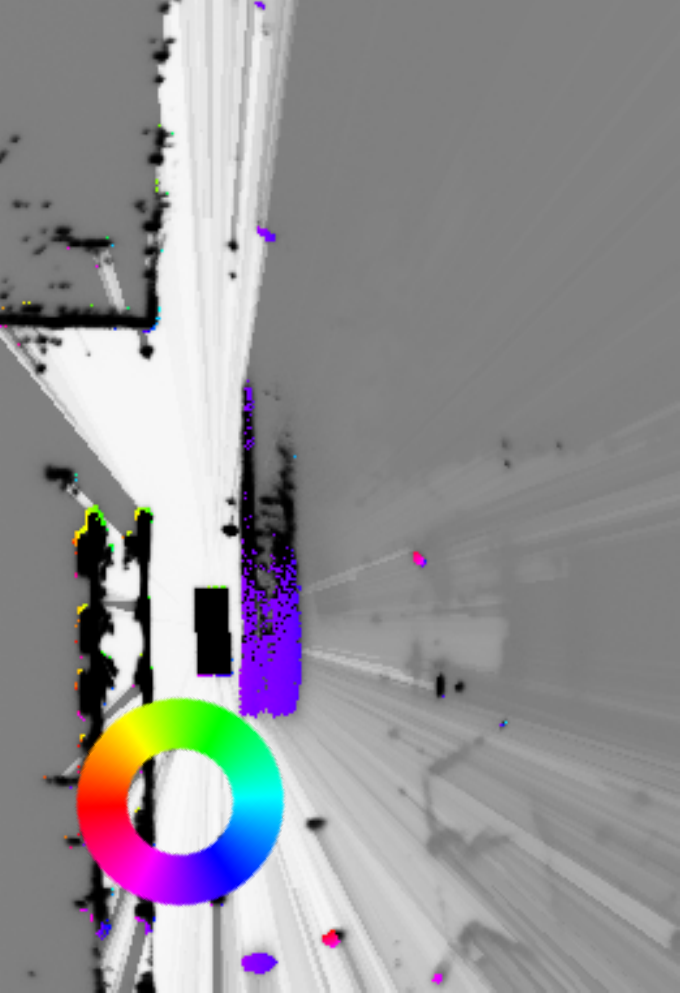}}%
    \put(0.01556439,0.02970319){\color[rgb]{0,0,0}\makebox(0,0)[lt]{\lineheight{1.25}\smash{\begin{tabular}[t]{l}\bfseries orientation\end{tabular}}}}%
    \put(0,0){\includegraphics[width=\unitlength,page=2]{in_615.pdf}}%
  \end{picture}%
\endgroup%

		\caption{t\textsubscript{0}}
	\end{subfigure}
	\qquad
	\begin{subfigure}{0.4\columnwidth}
		\centering
		\def\svgwidth{\columnwidth}		
		%% Creator: Inkscape inkscape 0.92.3, www.inkscape.org
%% PDF/EPS/PS + LaTeX output extension by Johan Engelen, 2010
%% Accompanies image file 'feed_615_15.pdf' (pdf, eps, ps)
%%
%% To include the image in your LaTeX document, write
%%   \input{<filename>.pdf_tex}
%%  instead of
%%   \includegraphics{<filename>.pdf}
%% To scale the image, write
%%   \def\svgwidth{<desired width>}
%%   \input{<filename>.pdf_tex}
%%  instead of
%%   \includegraphics[width=<desired width>]{<filename>.pdf}
%%
%% Images with a different path to the parent latex file can
%% be accessed with the `import' package (which may need to be
%% installed) using
%%   \usepackage{import}
%% in the preamble, and then including the image with
%%   \import{<path to file>}{<filename>.pdf_tex}
%% Alternatively, one can specify
%%   \graphicspath{{<path to file>/}}
%% 
%% For more information, please see info/svg-inkscape on CTAN:
%%   http://tug.ctan.org/tex-archive/info/svg-inkscape
%%
\begingroup%
  \makeatletter%
  \providecommand\color[2][]{%
    \errmessage{(Inkscape) Color is used for the text in Inkscape, but the package 'color.sty' is not loaded}%
    \renewcommand\color[2][]{}%
  }%
  \providecommand\transparent[1]{%
    \errmessage{(Inkscape) Transparency is used (non-zero) for the text in Inkscape, but the package 'transparent.sty' is not loaded}%
    \renewcommand\transparent[1]{}%
  }%
  \providecommand\rotatebox[2]{#2}%
  \newcommand*\fsize{\dimexpr\f@size pt\relax}%
  \newcommand*\lineheight[1]{\fontsize{\fsize}{#1\fsize}\selectfont}%
  \ifx\svgwidth\undefined%
    \setlength{\unitlength}{195.75bp}%
    \ifx\svgscale\undefined%
      \relax%
    \else%
      \setlength{\unitlength}{\unitlength * \real{\svgscale}}%
    \fi%
  \else%
    \setlength{\unitlength}{\svgwidth}%
  \fi%
  \global\let\svgwidth\undefined%
  \global\let\svgscale\undefined%
  \makeatother%
  \begin{picture}(1,1.45977011)%
    \lineheight{1}%
    \setlength\tabcolsep{0pt}%
    \put(0,0){\includegraphics[width=\unitlength,page=1]{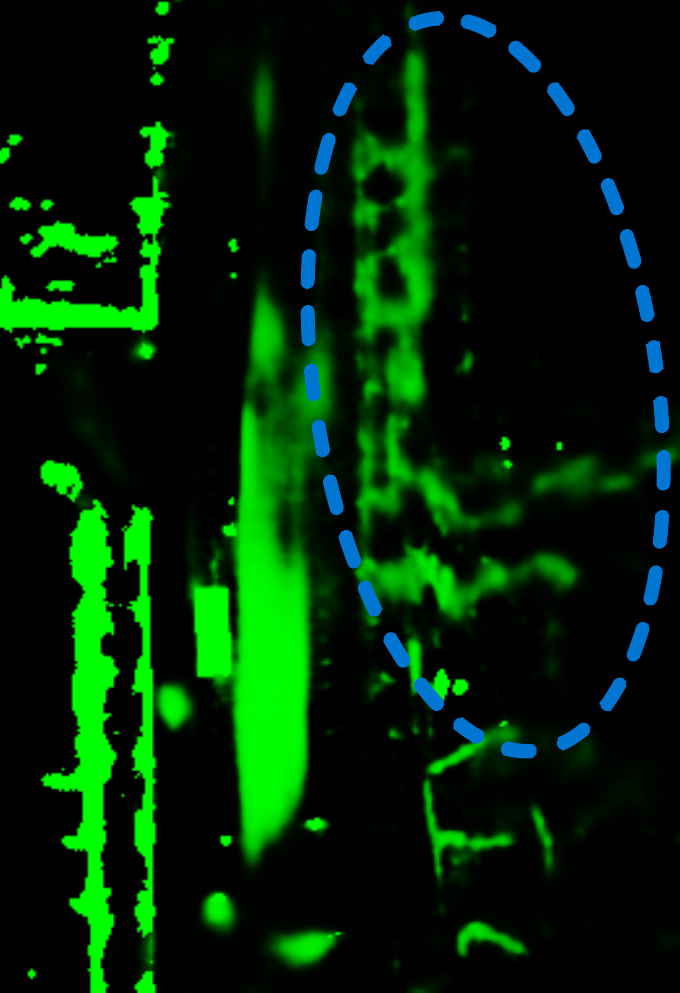}}%
    \put(0.02,0.02){\color[rgb]{1,1,1}\makebox(0,0)[lt]{\lineheight{1.25}\smash{\begin{tabular}[t]{l}\bfseries1.5\,s\end{tabular}}}}%
  \end{picture}%
\endgroup%

		\caption{Feedforward skips}
	\end{subfigure}
	%\hspace{0.6cm}
	\begin{subfigure}{0.4\columnwidth}	
		\centering
		\def\svgwidth{\columnwidth}		
		%% Creator: Inkscape inkscape 0.92.3, www.inkscape.org
%% PDF/EPS/PS + LaTeX output extension by Johan Engelen, 2010
%% Accompanies image file 'rec_615_15.pdf' (pdf, eps, ps)
%%
%% To include the image in your LaTeX document, write
%%   \input{<filename>.pdf_tex}
%%  instead of
%%   \includegraphics{<filename>.pdf}
%% To scale the image, write
%%   \def\svgwidth{<desired width>}
%%   \input{<filename>.pdf_tex}
%%  instead of
%%   \includegraphics[width=<desired width>]{<filename>.pdf}
%%
%% Images with a different path to the parent latex file can
%% be accessed with the `import' package (which may need to be
%% installed) using
%%   \usepackage{import}
%% in the preamble, and then including the image with
%%   \import{<path to file>}{<filename>.pdf_tex}
%% Alternatively, one can specify
%%   \graphicspath{{<path to file>/}}
%% 
%% For more information, please see info/svg-inkscape on CTAN:
%%   http://tug.ctan.org/tex-archive/info/svg-inkscape
%%
\begingroup%
  \makeatletter%
  \providecommand\color[2][]{%
    \errmessage{(Inkscape) Color is used for the text in Inkscape, but the package 'color.sty' is not loaded}%
    \renewcommand\color[2][]{}%
  }%
  \providecommand\transparent[1]{%
    \errmessage{(Inkscape) Transparency is used (non-zero) for the text in Inkscape, but the package 'transparent.sty' is not loaded}%
    \renewcommand\transparent[1]{}%
  }%
  \providecommand\rotatebox[2]{#2}%
  \newcommand*\fsize{\dimexpr\f@size pt\relax}%
  \newcommand*\lineheight[1]{\fontsize{\fsize}{#1\fsize}\selectfont}%
  \ifx\svgwidth\undefined%
    \setlength{\unitlength}{195.75bp}%
    \ifx\svgscale\undefined%
      \relax%
    \else%
      \setlength{\unitlength}{\unitlength * \real{\svgscale}}%
    \fi%
  \else%
    \setlength{\unitlength}{\svgwidth}%
  \fi%
  \global\let\svgwidth\undefined%
  \global\let\svgscale\undefined%
  \makeatother%
  \begin{picture}(1,1.45977011)%
    \lineheight{1}%
    \setlength\tabcolsep{0pt}%
    \put(0,0){\includegraphics[width=\unitlength,page=1]{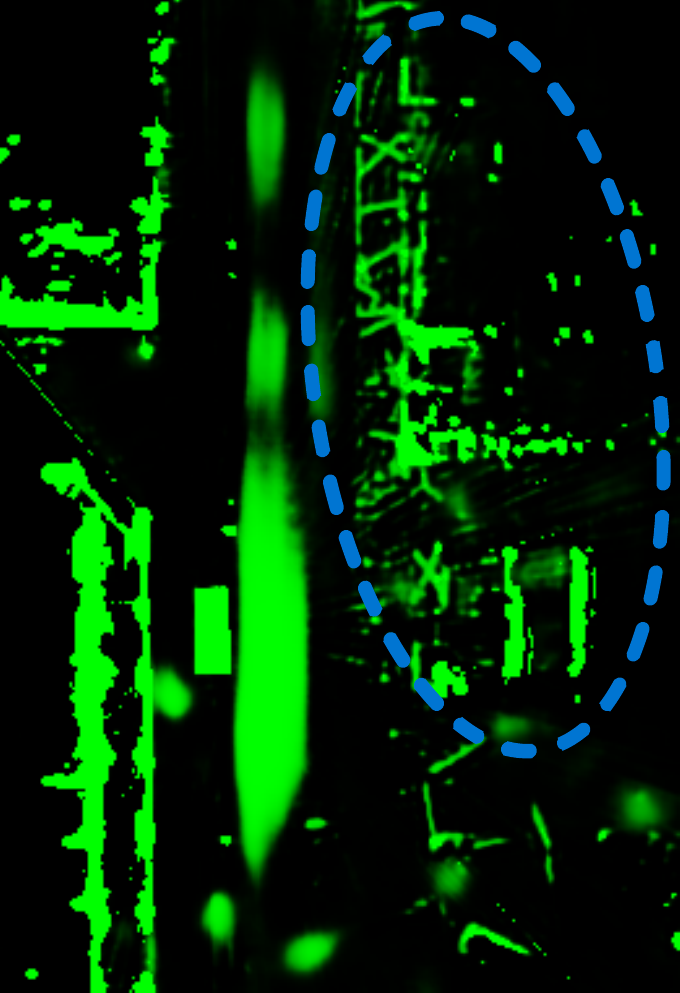}}%
    \put(0.02,0.02){\color[rgb]{1,1,1}\makebox(0,0)[lt]{\lineheight{1.25}\smash{\begin{tabular}[t]{l}\bfseries1.5\,s\end{tabular}}}}%
  \end{picture}%
\endgroup%

		\caption{Recurrent skips}
	\end{subfigure}
	\caption[Prediction of occluded static environment]{Illustration of the influence of recurrent skip connections in the prediction of occluded static environment. The blue circle marks the area that is occluded in the input grid map at the current time $t_0$. The left image visualizes the static environment in this area at an earlier time in the sequence. 
	The prediction results (two right images) show that only the network with the recurrent skip architecture recognizes the dense structures of the environment.
	}
	\label{fig:occ_static}
\end{figure*}
In addition, the recurrent skip connections are beneficial, in situations where the static environment is occluded. 
A skip architecture provides local information from shallow layers to get dense pixel wise predictions at the output. 
The ConvLSTMs in every skip connection enable the model to memorize this local information for static environment, even when there is occlusion.
In Fig. \ref{fig:occ_static} the predictions of the Grid Predictor Model with and without recurrent skip connections are compared in a situation with occluded static environment.
The left image shows the input grid map, where the static environment is unoccluded, at the current time $t_0$ a large vehicle causes occlusion of the static objects, marked with a blue circle.
The two images on the right show the outputs of the two network variants in green.
The Grid Predictor Model with the feedforward skip connections only produces blurry predictions for the occluded static environment. 
Here, the model has the semantic information from the ConvLSTMs in the deep layers, but misses the local information from shallow layers. 
In contrast, the Grid Predictor Model with recurrent skip connections produces dense predictions of the occluded static environment, because of the local information memorized by the recurrent skip connections.
\section{Conclusions}\label{sec:conclusions}
In this work we presented a recurrent deep learning approach to produce long-term predictions of the vehicle environment in a grid map representation using DOGMas as input data.
We developed a network architecture with ConvLSTMs in an Encoder-Decoder framework, yielding enhanced results with the use of less parameters compared to the previous work in \cite{DBLP:journals/corr/HoermannBD17}.
In addition, a novel recurrent skip architecture is developed which shows promising performance in dealing with missing input data as a result of occlusions in the DOGMa.

In this work, the data is only recorded by a stationary sensor, so in a following work the presented approach should be applied with recordings of a driving scenario. 
%
%\section{Conclusions}
%
%A conclusion section is not required. Although a conclusion may review the main points of the paper, do not replicate the abstract as the conclusion. A conclusion might elaborate on the importance of the work or suggest applications and extensions. 
%
%\addtolength{\textheight}{-12cm}   % This command serves to balance the column lengths
                                  % on the last page of the document manually. It shortens
                                  % the textheight of the last page by a suitable amount.
                                  % This command does not take effect until the next page
                                  % so it should come on the page before the last. Make
                                  % sure that you do not shorten the textheight too much.

%%%%%%%%%%%%%%%%%%%%%%%%%%%%%%%%%%%%%%%%%%%%%%%%%%%%%%%%%%%%%%%%%%%%%%%%%%%%%%%%

%%%%%%%%%%%%%%%%%%%%%%%%%%%%%%%%%%%%%%%%%%%%%%%%%%%%%%%%%%%%%%%%%%%%%%%%%%%%%%%%

%%%%%%%%%%%%%%%%%%%%%%%%%%%%%%%%%%%%%%%%%%%%%%%%%%%%%%%%%%%%%%%%%%%%%%%%%%%%%%%%
%\section*{APPENDIX}
%
%Appendixes should appear before the acknowledgment.
%
%\section*{ACKNOWLEDGMENT}
%
%The preferred spelling of the word ÒacknowledgmentÓ in America is without an ÒeÓ after the %ÒgÓ. Avoid the stilted expression, ÒOne of us (R. B. G.) thanks . . .Ó  Instead, try ÒR. B. %G. thanksÓ. Put sponsor acknowledgments in the unnumbered footnote on the first page.
%
%%%%%%%%%%%%%%%%%%%%%%%%%%%%%%%%%%%%%%%%%%%%%%%%%%%%%%%%%%%%%%%%%%%%%%%%%%%%%%%%
%
%\nocite{*} 
%
\bibliographystyle{IEEEtran}
\bibliography{IEEEtranControl,IEEEabrv,mybibfile}
\end{document}